\pdfoutput=1

\PassOptionsToPackage{sort}{natbib}
\documentclass[11pt]{article}

\usepackage[]{acl}

\usepackage{times}
\usepackage{latexsym}
\usepackage{booktabs}

\usepackage[T1]{fontenc}

\usepackage[utf8]{inputenc}

\usepackage{microtype}

\usepackage{xcolor}
\usepackage{array, makecell}
\usepackage[utf8]{inputenc}
\usepackage[inline]{enumitem}
\usepackage{adjustbox}
\usepackage{fancyvrb}
\usepackage{amsmath,amssymb,amsthm}
\usepackage{makecell}
\usepackage{svg}
\usepackage{bbm}
\usepackage{mathtools}
\usepackage{tabularx}
\usepackage{multirow}
\usepackage{pdfpages}
\usepackage{caption}
\usepackage{subcaption}
\usepackage{algpseudocode}
\usepackage{acronym}
\usepackage{hyperref}
\usepackage[most]{tcolorbox}
\usepackage{devanagari}
\usepackage{letltxmacro}
\usepackage{devanagaridigits}

\def\bng{\bngx}

\font\bngx=bang10

\def\*#1*#2{o\null{#2}{#1}}

\def\sh#1{\setbox0=\hbox{#1}%
     \kern-.02em\copy0\kern-\wd0
     \kern.04em\copy0\kern-\wd0
     \kern-.02em\raise.0433em\box0 }
\usepackage{array}
\newcommand{\headingnodot}[1]{\vspace*{1mm}\noindent\textbf{#1}}
\newcommand{\heading}[1]{\headingnodot{#1.}}

\acrodef{IR}{information retrieval}
\acrodef{LLM}{large language model}
\acrodef{NLP}{natural language processing}
\acrodef{NQ}{natural language question}
\acrodef{SQL2NQ}{SQL-to-NQ}
\acrodef{PEFT}{parameter-efficient fine-tuning}
\acrodef{tableQA}{table question answering}
\acrodef{ICL}{in-context learning}
\acrodef{BanglaTabQA}{Bengali Table Question Answering}
\acrodef{HindiTabQA}{Hindi Table Question Answering}

\LetLtxMacro\oldttfamily\ttfamily
\DeclareRobustCommand{\ttfamily}{\oldttfamily\csname ttsize\endcsname}
\newcommand{\setttsize}[1]{\def\ttsize{#1}}%
\setttsize{\small} 

\newtheorem{example}{Example}

\newtcolorbox[list inside=prompt,auto counter,number within=section]{llmPrompt}[1][]{
    colbacktitle=black!60,
    coltitle=white,
    fontupper=\footnotesize,
    boxsep=5pt,
    left=0pt,
    right=0pt,
    top=0pt,
    bottom=0pt,
    boxrule=1pt,
    #1,
}

\theoremstyle{definition}

\title{Table Question Answering for Low-resourced Indic Languages}

\author{
Vaishali Pal$^{1,2}$
\qquad
Evangelos Kanoulas$^1$
\qquad
Andrew Yates$^1$
\qquad
Maarten de Rijke$^1$
\\
$^1$University of Amsterdam, The Netherlands \\
\qquad
$^2$Discovery Lab, Elsevier, The Netherlands \\
\texttt{v.pal, e.kanoulas, a.c.yates, m.derijke@uva.nl}
}

\begin{document}
\maketitle
\begin{abstract}
  TableQA is the task of answering questions over tables of structured information, returning individual cells or tables as output. TableQA research has focused primarily on high-resource languages, leaving medium- and low-resource languages with little progress due to scarcity of annotated data and neural models. We address this gap by introducing a fully automatic large-scale \ac{tableQA} data generation process for low-resource languages with limited budget. We incorporate our data generation method on two Indic languages, Bengali and Hindi, which have no tableQA datasets or models. TableQA models trained on our large-scale datasets outperform state-of-the-art LLMs. We further study the trained models on different aspects, including mathematical reasoning capabilities and zero-shot cross-lingual transfer. Our work is the first on low-resource tableQA focusing on scalable data generation and evaluation procedures. Our proposed data generation method can be applied to any low-resource language with a web presence. We release datasets, models, and code.\footnote{\url{https://github.com/kolk/Low-Resource-TableQA-Indic-languages}}
\end{abstract}

\section{Introduction}
Tables are ubiquitous for storing information across domains and data sources such as relational databases, web articles, Wikipedia pages, etc.~\citep{10.1145/3404835.3462806}. Tables introduce new challenges in machine comprehension not present in text as they are are not well-formed sentences but a semi-structured collection of facts (numbers, long-tail named entities, etc.)~\citep{Jauhar2016TablesAS,iyyer-etal-2017-search,zhu2021tat,liu2021tapex,aitqa2022,10.1162/tacl_a_00446,pal-etal-2022-parameter,https://doi.org/10.48550/arxiv.2207.05270}.
Additionally, tables make position (rows/columns) bias~\cite{lin-etal-2023-inner} and entity popularity bias~\cite{gupta-etal-2023-temptabqa} severe. 
The \ac{tableQA} task introduces novel challenges compared to text-based question answering (text\-QA)~\cite{herzig-etal-2020-tapas,liu2021tapex,yu-etal-2018-spider,zhao-etal-2022-multihiertt,10.1145/3539618.3591708}. In addition to the semi-structured nature of tables, a tabular context leads to a high frequency of fact-based questions, mathematical and logical operations such as arithmetic~\cite{zhu2021tat}, set, relational~\cite{liu2021tapex,jiang-etal-2022-omnitab}, and table operations such as table joins~\cite{pal2023multitabqa}. Effective \ac{tableQA} systems not only have machine comprehension skills, but also numeracy understanding \citep{zhu2021tat,liu2021tapex,zhao-etal-2022-multihiertt,cheng-etal-2022-fortap}, table reasoning \cite{liu2021tapex,yu-etal-2018-spider}, table summarization~\cite{zhao-etal-2023-qtsumm,zhang2024qfmtsgeneratingqueryfocusedsummaries} and answer table generation ability~\cite{pal2023multitabqa}. 

Low-resource \ac{tableQA} aims to answer questions over semi-structured tables storing cultural and region-specific facts in a low-resource language. \citet{joshi-etal-2020-state} show that most languages struggle to be represented and are deprived of advances in NLP research. As manual data collection is slow and expensive, low-resource languages struggle with large-scale, annotated data for effective transfer learning solutions. The low-resource setting \cite{hedderich-etal-2021-survey,rudder-4-major-problems}  exacerbates the challenges of \ac{tableQA} with challenges of data sparsity, annotated data costs, and lack of trained models. In contrast to textQA, syntactico-semantic variations such as agreement and morphology are not exhibited in tables, but high presence of culturally significant yet long-tail entities makes adapting existing high resource datasets and trained models challenging. Research on low-resource table inference \cite{minhas-etal-2022-xinfotabs} shows that standard approaches of translating English datasets for low-resource data creation are infeasible for tables due to high translation error as tables are not well-formed sentences. 

\heading{Challenges}
Our work focuses on studying the following core challenges of low-resource \ac{tableQA}:
\begin{enumerate}[label=(\arabic*),leftmargin=*,nosep]
    \item low-resource \textbf{tableQA data scarcity} and under-representation of cultural facts.
    \item Existing \textbf{neural models' poor alignment} in low-resource languages and a lack of understanding of table structure. 
\end{enumerate}
\noindent
This motivates us to explore low-resource \ac{tableQA} by designing a low-cost and large-scale automatic data generation and quality estimation pipeline. We discuss the process in detail with a low-resource Indic language, Bengali (spoken extensively in Bangladesh and India, with over 230 million native speakers~\cite{DeepHateExplainer}), and explore generalizability  with Hindi (570 million speakers).

Our main contributions are as follows:
\begin{enumerate}[label=(\arabic*),leftmargin=*,nosep]
    \item We introduce \textbf{low-resource tableQA task}.
    \item  We design a \textbf{method} for automatically generating low-resource \ac{tableQA} data in a scalable  budget-constrained manner.
    \item We release \textbf{resources} to support low-resource \ac{tableQA}: Large-scale \ac{tableQA} \textbf{datasets} and \textbf{models} for 2 Indic languages, Bengali (\ac{BanglaTabQA}) and Hindi (\ac{HindiTabQA}). BanglaTabQA contains $19K$ Wikipedia tables, $2M$ training, $2K$ validation and  $165$ test samples. HindiTabQA contains $2K$ Wikipedia tables, $643K$ training, $645$ validation and $125$ test samples. 
\end{enumerate}

\section{Related Work}
TableQA aims to answer a user question from semi-structured input tables. Prior work on \ac{tableQA} in English can be classified as extractive \cite{herzig-etal-2020-tapas,tabert2020} or abstractive \cite{10.1162/tacl_a_00446,pal-etal-2022-parameter,zhao2023qtsumm,10.1145/3539618.3591708}. 
While extractive \ac{tableQA} focuses on row and cell selection \cite{herzig-etal-2020-tapas}, abstractive \ac{tableQA} generates various types of answers such as factoid answers \cite{liu2021tapex}, summaries \cite{zhao2023qtsumm,zhang2024qfmtsgeneratingqueryfocusedsummaries}, or answer tables \cite{pal2023multitabqa}. Low-resource setting poses challenges for various NLP tasks. The low-resource corpus creation \cite{hasan-etal-2020-low,10.1007/s11042-021-11228-w,bhattacharjee-etal-2022-banglabert} has used automatic annotation efforts by synthesizing a large-scale dataset. \citet{10.1007/s11042-021-11228-w} train a Bengali QA system by developing a synthetic dataset translated from standard English QA datasets. \citet{bhattacharjee-etal-2022-banglabert,hasan-etal-2020-low} create low-resource datasets by translating English datasets to Bengali using neural models. However, these methods are unsuitable due to the semi-structured ungrammatical sequential representation of tables. 


\section{Task Definition}

We formulate low-resource \ac{tableQA} as a sequence generation task. Given a question $Q$ of $k$ tokens $q_1, q_2, \ldots , q_k$, and table $T$ comprising of $m$ rows and $n$ columns $\{h_1, \ldots ,h_n$, $t_{1,1}$, $t_{1,2}, \ldots, t_{1,n}, \ldots, t_{m,1}$, $t_{m,2}, \ldots, t_{m,n}\}$ where $t_{i,j}$ is value of the cell at the $i$-th row and $j$-th column and $h_j$ is the $j$-th column header; the low-resource \ac{tableQA} model generates an answer table $T_{out}$. The input sequence is the concatenated question $Q$, and linearized input table $T$ separated by special sentinel tokens. The answer, $T_{out}$, is also a linearized sequence. Henceforth, for concreteness, we will use Bengali as the example low-resource language. 
The input to such a model is:
\begin{quote}
\vspace{-0.5em}
\small
$q_1\ q_2\ldots q_k$ <{\bng klam}> $h_i\ldots h_n$ <{\bng \*r*ea 1}> $t_{1,1}\ldots t_{1,n}$ <{\bng \*r*ea} i> $ t_{i,j}\ldots  t_{i,n}\ldots$ <{\bng \*r*ea} m> $t_{m,1}\ldots t_{m,n}$. 
\end{quote}
\vspace{-0.5em}
The answer table, $T_{out}$, is a linearized sequence:
\vspace{-0.5em}
\begin{quote}
\small
<{\bng klam}> $H_i\ldots H_q$ <{\bng \*r*ea 1}> $o_{1,1}\ldots o_{1,q}$ <{\bng \*r*ea} i> $ o_{i,j}\ldots  o_{i,q}\ldots$ <{\bng \*r*ea} m> $o_{p,1}\ldots o_{p,q}$
\end{quote}
\vspace{-0.5em}
where $o_{i,j}$ is value at the $i$-th row and $j$-th column and $H_j$ is the $j$-th column header of $T_{out}$.


\section{Methodology for Dataset Generation}
\label{sec:dataset_method}

\begin{figure*}[t!]
 \centering
 \includegraphics[width=0.9\textwidth]{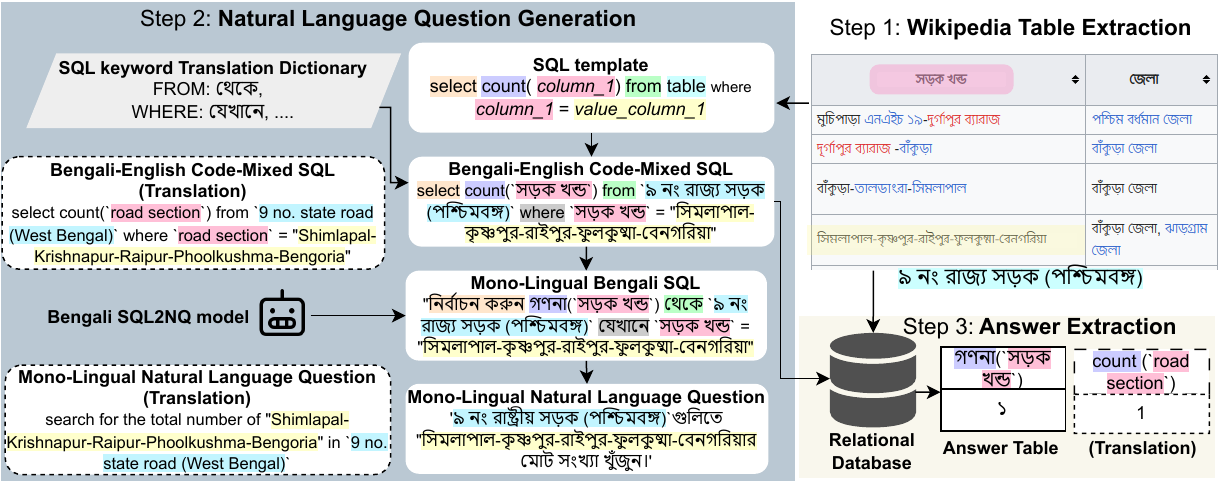}
 \caption{\textbf{BanglaTabQA Dataset generation}: The SQL elements and table elements are color-coordinated to represent a single SQL/table element. Dotted rectangles represent translations for accessibility to non-native readers.}
 \label{fig:data_creation_bengali}
\end{figure*}

Effective training of low-resourced \ac{tableQA} requires creation of large-scale datasets of questions, input and answers tables, to align a language model to the low-resource language and adapt it to semi-structured tables and QA task. We address \textbf{Challenge 1} by designing an automatic data generation process to generate a large-scale low resource \ac{tableQA} corpus of training and validation samples. We follow a 3-step pipeline as follows: 
\begin{enumerate*}[label=(\roman*)]
    \item table extraction,
    \item question generation, and 
    \item answer table extraction.
\end{enumerate*}
This pipeline applied on Bengali, as depicted in Figure~\ref{fig:data_creation_bengali}, generates the \textbf{\ac{BanglaTabQA}} dataset. 

\subsection{Table Extraction}
English Wikipedia with $6,751,000+$ articles is used for English \ac{tableQA} datasets \cite{pasupat-liang-2015-compositional}, but is insufficient for non-Latin languages with many cultural topics missing. The standard process~\cite{10.1007/s11042-021-11228-w,bhattacharjee-etal-2022-banglabert} of translating English datasets to low-resource languages is biased due to lack of cultural topic/fact representation in English \ac{tableQA} datasets. For example, the named-entity {\bng Adhiraj gaNG/guil} (Adhiraj Ganguly), exists only in Bengali Wikipedia,\footnote{\href{https://bn.wikipedia.org/s/er3k}{https://bn.wikipedia.org/wiki/{\bng Adhiraj}\_{\bng gaNG/guil}}} and not in English. Further, translating English tables with machine translation models is error-prone \cite{minhas-etal-2022-xinfotabs} as tables are not well-formed sentences but collections of facts. To mitigate these issues, we extract tables from Wikipedia dump of the low-resource language. 

\subsection{Natural Language Question Generation}
\label{sec:nq_generation}
The question generation is a 2-step process: 

\paragraph{Code-mixed SQL query generation.} 
\label{sec:code_mixed_sql_gen}
We automatically generate SQL queries over the extracted low-resourced tables with SQL templates from the SQUALL dataset \cite{Shi:Zhao:Boyd-Graber:Daume-III:Lee-2020}. These templates have placeholders of table components such as table name, column names, etc. which are randomly assigned with values from a Wikipedia table. For example, the template ``\texttt{select count(c1) from w where c1 = value}'' is instantiated by assigning a Bengali table name ``{\bng 9 noNNG rajYo sorhok (pish/cm bNG/g)}'' to \texttt{w}, column header ``{\bng ejla}'' to \texttt{c1}, and ``{\bng baNNkurha ejla}'' to \texttt{value}. This results in an executable code-mixed query ``\texttt{select count({\bng ejla}) \texttt{from} {\bng 9 noNNG rajYo sorhok (pish/cm bNG/g)} where `{\bng ejla}` = "{\bng baNNkurha ejla}"}'', where the SQL keywords are in English but all table information is in the low-resource language (Bengali). This leads to $13,345,000$ executable Bengali code-mixed queries.

\paragraph{Natural language question generation.} 
\label{para:nq_gen}
We formulate question generation as a sequence-to-sequence task by transforming a code-mixed SQL query into a \ac{NQ}. To the best of our knowledge, there exists no sequence generation models which translates code-mixed SQL queries to low-resource natural language questions. To train a model for this conversion, we first transform the code-mixed SQL to a monolingual SQL-like query in the low-resource language. As the only linguistic variation exhibited in the SQL templates is polysemy i.e. a dearth of one-to-one correspondence between English SQL keywords and the corresponding low-resource language translations, we employ native speakers well-versed in SQL to manually create one-to-one mappings of $27$ SQL keywords for linguistic transfer of SQL keywords to the corresponding low-resource language. All table-specific words are directly copied into the monolingual query. We discard \texttt{FROM} keyword and table name from the query as it is associated with a single input table. This leads to a SQL-like monolingual query in the low-resource language which is a well-formed sentence. For example, code-mixed Bengali query ``\texttt{\textcolor{teal}{select} \textcolor{orange}{count}(  \textcolor{violet}{`{\bng ejla}`}) \textcolor{purple}{from} \textcolor{olive}{\bng{9 noNNG rajYo sorhok (pish/cm bNG/g)}}  \textcolor{blue}{where} \textcolor{brown}{`{\bng ejla}` = "{\bng baNNkurha ejla}"}}'', results in a monolingual Bengali query ``\textcolor{teal}{{\bng inr/bacon korun}} \textcolor{orange}{\bng gNna}(
\textcolor{violet}{`{\bng ejla}`}) 
 \textcolor{blue}{{\bng eJkhaen}} \textcolor{brown}{`{\bng ejla}` = "{\bng baNNkurha ejla}"}''. In contrast to tables which are invalid sentences, queries and \ac{NQ} are well-formed sequences and effectively transformed (SQL to question) with existing encoder-decoder models.  We train a \ac{SQL2NQ} model (\texttt{mbart-50-large}~\cite{liu-etal-2020-multilingual-denoising} backbone) by translating $68,512$ training and $9,996$ validation samples from semantic parsing datasets: Spider \cite{yu-etal-2018-spider}, WikiSQL \cite{zhong2017seq2sql}, Atis \cite{price-1990-evaluation,dahl-etal-1994-expanding}, and Geoquery \cite{10.5555/1864519.1864543} to the low-resource language. We use this \Ac{SQL2NQ} model to transform the queries to \Ac{NQ}. For example, Bengali \ac{SQL2NQ} model transforms the aforementioned query to the \ac{NQ} ``{\bng \textcolor{orange}{kbar} \textcolor{brown}{baNNkurha} \textcolor{violet}{ejlar} \textcolor{teal}{Uel/lk Aaech}?}''.

\subsection{Answer Table Extraction} 
\label{sec:answer_table_extraction}
We dump low-resource Wikipedia tables in a relation database. The code-mixed SQL queries are executed with an SQL compiler over a relational database comprising of the low-resourced Wikipedia tables to extract the answer tables. We execute the $13,345,000$ Bengali code-mixed queries to extract the corresponding answer tables.

\subsection{Automatic Quality Control} 
\label{sec:quality_control}
We employ automatic quality control steps to ensure quality of the synthetic \ac{tableQA} data. 

\paragraph{Code-mixed query and answer quality control.} We discard all code-mixed queries which execute to an error with an SQL compiler. This process follows the quality control in \cite{pal2023multitabqa} and discards invalid and erroneous queries and samples. 

\begin{figure}[t]
 \centering
\includegraphics[width=\columnwidth]{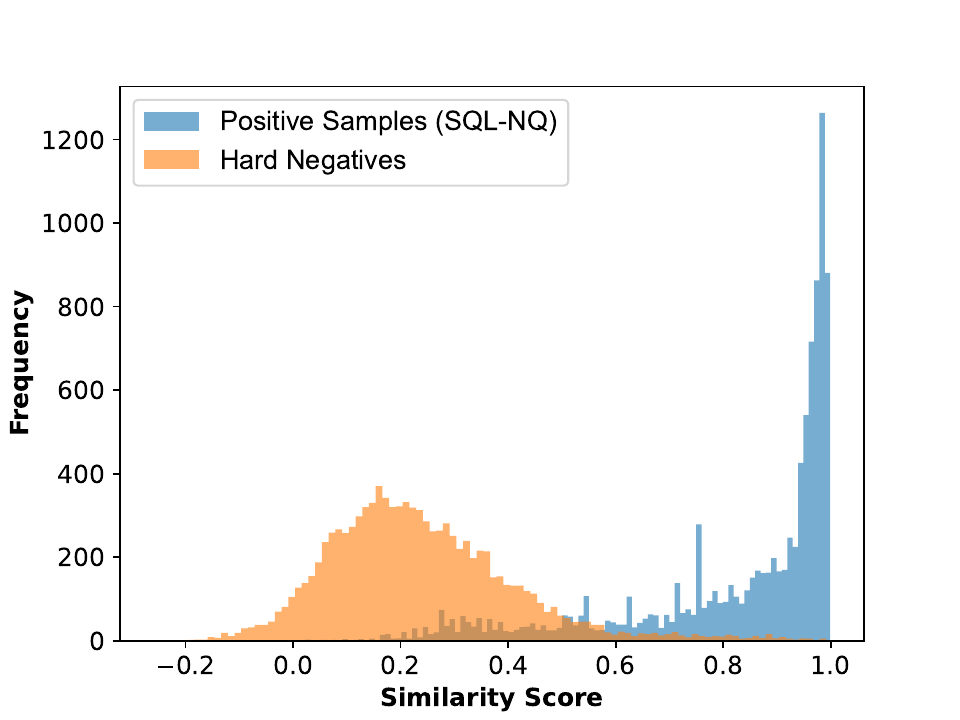}
    \caption{Histogram of similarity scores from fine-tuned Bengali SQL2NQSim model of $1,000$ random samples}
 \label{fig:labse_scores_after_finetuning}
\end{figure}

\paragraph{Natural Language Question quality control.} 
\label{sec:nq_quality_control}
We evaluate the quality of the generated \ac{NQ} with a sentence similarity model to discard questions that have low similarity score with the corresponding monolingual queries. We found the standard method of quality evaluation in low-resource languages~\cite{bhattacharjee-etal-2022-banglabert,ramesh-etal-2022-samanantar} using the sentence similarity model, LaBse \cite{2022_labse}, incompatible for code-mixed SQL-\ac{NQ} due to low discriminating ability ($0.55$ mean similarity score and $0.13$ standard deviation for Bengali SQL-NQ). For example, LaBse assigns low score ($0.43$) for positive SQL-NQ pair corresponding to the Bengali query ``\texttt{SELECT title ORDER BY year DESC LIMIT 1}" and Bengali \ac{NQ} ``\texttt{Return the most recent title corresponding to the most recent year}" (translated for non-native readers), while it assigns a high score ($0.8$) to negative pair ``\texttt{SELECT count(*) WHERE `work` = The World of Saudamini}" and the unrelated NQ ``\texttt{How many games scored a total of 4?}". Table~\ref{tab:quality_control_examples} in Appendix \ref{sec:labse_sql2nq_examples}  shows more examples.  This necessitates fine-tuning LaBse on low-resourced SQL-NQ samples. First, we use the translated semantic parsing samples ($68,512$ training and $9,996$ SQL-NQ pairs), described in Section \ref{para:nq_gen}, as positive pairs and in-batch negatives with multiple-negatives ranking loss. We call this the SQL2NQSim model. We select the best checkpoint by evaluating SQL2NQSim on $1,000$ randomly selected hard-negatives (unrelated/negative  SQL-negative question pairs for which pre-trained LaBse assigns a high similarity score ($> 0.5$)). We use that checkpoint to obtain similarity scores of the low-resourced \ac{tableQA} SQL-NQ pairs and discard samples with a similarity score lower than a threshold. We select a good threshold by plotting a histogram of scores assigned by the SQL2NQSim model on $10,000$ randomly selected positives and hard-negatives and selecting the inflection point as the threshold. Figure~\ref{fig:labse_scores_after_finetuning} shows the scores' histogram for \ac{BanglaTabQA}. 
We select a strict threshold of $0.74$ (hard-negatives scores taper-off around $0.7$). The final \ac{BanglaTabQA} dataset, after quality control, comprises of $2,050,296$ training and $2,053$ validation samples.

\subsection{Dataset Analysis}
In contrast to textQA, \ac{tableQA} focuses on mathematical questions~\cite{liu2021tapex,zhu2021tat,pal2023multitabqa}. Following \cite{liu2021tapex}, we analyse \ac{BanglaTabQA} dataset on question complexity, which estimates the difficulty of a question based on the corresponding SQL query. As \ac{tableQA} enforces mathematical, logical and table reasoning questions, we further classify \ac{tableQA} queries into different classes of table operations determined by the SQL operators present. 

\begin{figure}[h!]
 \centering
\includegraphics[width=\columnwidth]{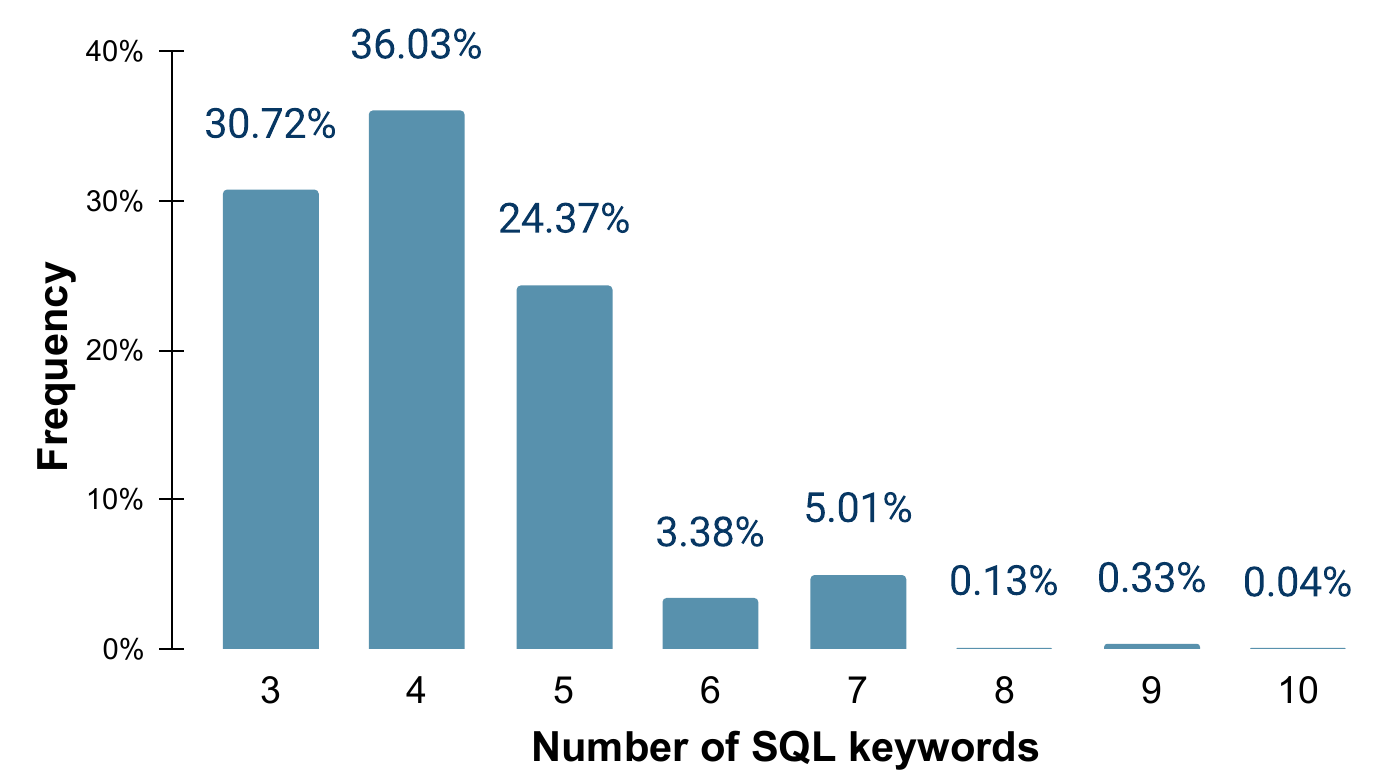}
    \caption{Number of SQL keywords per query histogram in the \ac{BanglaTabQA} dataset.
    }
 \label{fig:keyword_count}
\end{figure}

\begin{figure}[h!]
    \centering
    \includegraphics[width=\columnwidth]{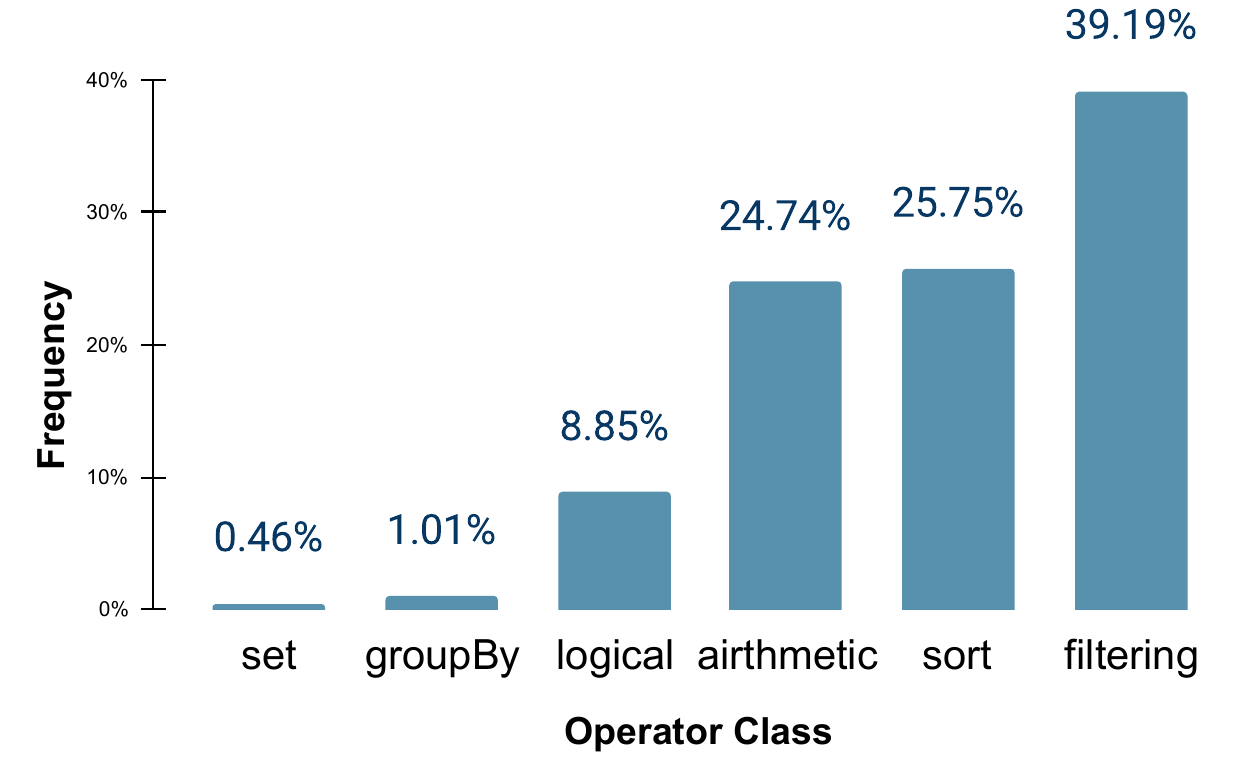}
    \caption{Histogram of operator classes in the BanglaTabQA dataset.
    }
 \label{fig:operator_class}
\end{figure}

\paragraph{Question complexity.} 
Recent work on \ac{tableQA} \cite{liu2021tapex} categorizes SQL queries into difficulty levels based on the number of SQL keywords. We follow this approach and count the number of keywords for each query. Figure \ref{fig:keyword_count} shows that most of \ac{BanglaTabQA} queries have $4$ SQL keywords. The longest SQL queries are comprised of 10 keywords, and the shortest ones of $3$ SQL keywords. 
      
\paragraph{Mathematical operations.} 
\label{sec:operator_classes} 
We further categorize each sample based on the operators present in the question. We utilize the SQL query associated with a question to extract all keywords for classification. We categorize data samples into $6$ operator classes: arithmetic, sorting, group by, filtering, set operators, and logical operators. Arithmetic operators comprises of SQL numeric operations such as \texttt{sum}, \texttt{count}, \texttt{min}, etc. Sorting refers to ordering of the answer values in an ascending or descending order. Group by is the SQL operator of grouping rows based on a criterion. Filtering corresponds to SQL operators such as \texttt{where} and \texttt{having} used to filter the input table. Set operators involve \texttt{union}, \texttt{intersect}, and \texttt{except}. Finally, we classify logical operators to be conjunction (\texttt{and}) and disjunction (\texttt{or}) to combine filtering conditions. It also includes membership operators (\texttt{in}, \texttt{between}, etc.) and string matching operator (\texttt{like}). The classification of the operators is shown in Table~\ref{tab:operator_class}.  Figure \ref{fig:operator_class} shows the distribution of the 6 operator classes for the \ac{BanglaTabQA} dataset. 

\subsection{Test Set} 
\label{sec:test_set}
We manually annotate test samples for evaluating low-resource \ac{tableQA} models on clean data. We select unique tables not present in the training and validation set to avoid data leakage. To ensure question diversity, we select code-mixed SQL  representing each of the 6 operator classes (discussed in Section \ref{sec:operator_classes}) and distinct from the training and validation data. Three native annotators well-versed in SQL were employed for annotation. One annotator was tasked with question generation and given the synthetic SQL query, input tables and the answer table, and asked to rewrite the code-mixed query to a natural language question. The remaining two were tasked with evaluation of the question generated by the first annotator. The evaluator-annotators were provided the code-mixed query, input table, answer table, and the annotated question and asked to rate the question based on fluency. We estimate the annotated question fluency with a 5-point Likert scale (1-5), where a higher score indicates a better fluency. The final score for each question was computed by averaging the scores of the evaluator-annotators. For \ac{BanglaTabQA}, we manually annotate $165$ test samples. We estimate an inter-annotator agreement with Fliess's Kappa score \cite{Fleiss1971MeasuringNS} of $0.82$, indicating strong agreement among the annotators. The average fluency score across test set questions was $4.3$, indicating high fluency.

\subsection{Generalizability of Dataset Methodology}
\label{sec:HindiTableQA}
We study the generalizability of the dataset generation method by repeating the process on another Indic language: Hindi (Hi) with more than 602 million speakers. To the best of our knowledge, there is no existing \ac{tableQA} data for Indic languages. Hindi text is in Devanagari script which is different from Bengali written in Eastern-Nagari (Bengali-Assamese) script. This requires \ac{tableQA} models to be trained on large-scale Hindi datasets for good alignment. Following the dataset creation process in Section~\ref{sec:dataset_method}, we extract $1,921$ Hindi tables from the respective Wikipedia dumps. We generate $82,00,000$ Hindi code-mixed queries automatically to extract answer tables and generate the Hindi natural language questions. The final HindiTabQA dataset comprises of $643,434$ synthetic training, $645$ synthetic validation samples and $121$ manually annotated test samples.

\section{Experimental Setup}
\label{sec:experimental_setup}
We address \textbf{Challenge 2} by studying the effectiveness of state-of-the-art models (baselines) in \textit{Bengali table QA}. Experimental results (Section \ref{sec:results}) show the need for a large-scale \ac{BanglaTabQA} dataset and model training. We analyze several models' effectiveness in Bengali language, mathematical/table operations and generalizability, thus providing a measure of the dataset quality and consequently the dataset creation methodology.

\paragraph{Baselines.} 
We perform 2-shot \ac{ICL} to adapt \ac{LLM}s to \ac{BanglaTabQA} task. We further fine-tune an encoder-decoder model. The demonstrations are the concatenated question and flattened input table with the flattened answer table. We use the following models as baselines:

\begin{enumerate}[label=(\arabic*),leftmargin=*,nosep]
    \item \textbf{En2Bn:} We fine-tune an encoder-decoder model, \texttt{mbart-50-large}, with $25,000$ random samples from MultiTabQA's~\cite{pal2023multitabqa} pre-training data translated to Bengali using Google translate. MultiTabQA used SQUALL templates to generate their queries and have the same distribution as \ac{BanglaTabQA} queries. However, the input tables of MultiTabQA are English wiki-tables from WikiTableQuestions dataset \cite{pasupat-liang-2015-compositional} and are not representative of Bengali cultural topics/facts. 

    \item \textbf{OdiaG~\cite{OdiaGenAI-Bengali-LLM}} is Llama-7b~\cite{touvron2023llama} adapter-tuned (LoRA~\cite{hu2022lora}) on $252$k Bengali instruction set.\footnote{OdiaGenAI/odiagenAI-bengali-lora-model-v1}

    
    \item \textbf{GPT:} \texttt{GPT-3.5}~\cite{NEURIPS2020_1457c0d6} performs well on English \ac{tableQA} \cite{zha2023tablegpt}. \texttt{GPT-4}~\cite{openai2023gpt4} outperforms other \ac{LLM}s (Chinchilla~\cite{hoffmann2022training}, PaLM~\cite{chowdhery2022palm}) in low-resource languages, including Bengali and Hindi, on various tasks ($14,000$ multiple-choice problems on $57$ subjects in a translated MMLU benchmark~\cite{hendryckstest2021}).    
\end{enumerate}

\newcommand{\PreserveBackslash}[1]{\let\temp=\\#1\let\\=\temp}
\newcolumntype{C}[1]{>{\PreserveBackslash\centering}p{#1}}
\newcolumntype{R}[1]{>{\PreserveBackslash\raggedleft}p{#1}}
\newcolumntype{L}[1]{>{\PreserveBackslash\raggedright}p{#1}}

\begin{table*}[!t]
\centering
\small
\setlength{\tabcolsep}{2mm}
\begin{tabular}{@{}p{12mm}C{5mm}C{5mm}C{5mm}C{5mm} C{5mm}C{5mm}C{5mm}C{6mm} |C{5mm}C{5mm}C{5mm}C{5mm} C{5mm}C{5mm}C{5mm}C{5mm}}
\toprule
\multirow{3}{*}{\textbf{Model}} & \multicolumn{8}{c}{Bengali} & \multicolumn{8}{|c}{Hindi}\\
& \multicolumn{4}{c}{Validation Set scores (\%)} & \multicolumn{4}{c}{Test Set scores (\%)} &  \multicolumn{4}{|c}{Validation Set scores (\%)} & \multicolumn{4}{c}{Test Set scores (\%)}\\ 
\cmidrule(r){2-5}
\cmidrule(r){6-9}
\cmidrule(r){10-13}
\cmidrule{14-17}
 & \textbf{Tab} & \textbf{Row} & \textbf{Col} & \textbf{Cell}  & \textbf{Tab} & \textbf{Row} & \textbf{Col} & \textbf{Cell}  & \textbf{Tab} & \textbf{Row} & \textbf{Col} & \textbf{Cell}  & \textbf{Tab} & \textbf{Row} & \textbf{Col} & \textbf{Cell} \\
\midrule
 En2(Bn/Hi)  &  $\phantom{0}0.05$ & $\phantom{0}3.06$ & $\phantom{0}0.20$ & $\phantom{0}3.07$ & $\phantom{0}0.00$ & $\phantom{0}4.73$ & $\phantom{0}0.00$ & $\phantom{0}4.73$ & $\phantom{0}0.00$ & $\phantom{0}3.37$ & $\phantom{0}0.47$ & $\phantom{0}3.43$ & $\phantom{0}0.00$ & $\phantom{0}5.03$ & $\phantom{0}8.26$ & $\phantom{0}5.03$
 \\
 OdiaG  & $\phantom{0}0.00$ & $\phantom{0}3.89$ & $\phantom{0}0.00$ & $\phantom{0}3.89$ & $\phantom{0}0.69$ & $\phantom{0}1.77$ & $\phantom{0}0.69$ & $\phantom{0}1.4$2 & $-$ & $-$ & $-$  & $-$  & $-$  & $-$  & $-$  & $-$ 
 \\
 OpHathi & $-$ & $-$ & $-$ & $-$ & $-$ & $-$ & $-$ & $-$ & $\phantom{0}0.00$ & $\phantom{0}0.00$ & $\phantom{0}0.00$ & $\phantom{0}0.00$ & $\phantom{0}0.00$ & $\phantom{0}0.11$ & $\phantom{0}0.37$ & $\phantom{0}0.74$ 
 \\
 GPT-3.5 & $\phantom{0}1.14$ & $\phantom{0}4.81$ & $\phantom{0}1.67$ & $\phantom{0}5.14$ & $\phantom{0}6.04$ & $10.06$ & $\phantom{0}9.12$ & $\phantom{0}9.84$ & $\phantom{0}4.81$ & $\phantom{0}8.94$ & $\phantom{0}4.99$ & $\phantom{0}9.71$ & $\phantom{0}8.20$ & $10.29$ & $\phantom{0}7.10$ & $\phantom{0}9.81$\\
 GPT-4 & $\phantom{0}0.00$ & $13.57$ & $\phantom{0}5.43$ & $14.65$ & $26.83$ & $\textbf{38.67}$ & $26.74$ & $\textbf{36.51}$ &  $15.53$ & $22.60$ & $16.02$ & $22.25$ & $11.11$ & $21.49$ & $11.76$ & $20.84$ \\
 \midrule
 & 
 \multicolumn{8}{c|}{\textbf{BnTQA}} & \multicolumn{8}{c}{\textbf{HiTQA}} 
 \\
 \textbf{-llama} & $\textbf{60.08}$ & $\textbf{68.30}$ & $\textbf{60.47}$ & $\textbf{68.30}$ & $\phantom{0}9.41$ & $12.35$ & $9.85$ & $11.87$  & $14.76$ & $\phantom{0}9.92$ & $14.13$ & $\phantom{0}7.29$ & $13.11$ & $\phantom{0}9.71$ & $11.11$ & $\phantom{0}7.66$ 
 \\
 \textbf{-mBart} & $56.63$  & $64.10$ & $56.79$ & $64.31$ & $\textbf{35.88}$ & $33.16$ & $\textbf{35.88}$ & $33.16$ & $92.09$ & $87.97$ & $92.02$ & $87.97$ & $33.06$ & $43.35$ & $33.88$ & $43.35$ 
 \\
 \textbf{-M2M} & $45.31$ & $58.07$ & $45.29$ & $58.04$ & $28.05$ & $34.55$ & $28.05$ & $34.55$ & $89.55$ & $85.32$ & $89.34$ & $85.15$ & $28.93$ & $33.11$ & $28.92$ & $33.10$ 
 \\ 
 \textbf{-BnTQA} & $-$ & $-$ & $-$ & $-$ &  $-$ & $-$ & $-$ & $-$ & $92.40$ & $88.10$ & $92.42$ & $88.12$ & $\textbf{41.32}$ & $\textbf{47.26}$ & $\textbf{41.32}$ & $\textbf{47.26}$ 
 \\ 
\bottomrule
\end{tabular}
\caption{Baseline, BnTQA-X and HiTQA-X models' scores. -X represents the backbone architecture of a fine-tuned model and  $-$ entries are for incompatible models in a low-resourced language (Bengali or Hindi).}
\label{tab:results_all_lang}
\end{table*}
\paragraph{BanglaTabQA models.} 

Bengali \ac{tableQA} models must understand both Bengali \textbf{script} \emph{and} \textbf{numerals}, crucial for mathematical operations. However, Bengali numbers are not present in many state-of-the-art Indic models' \cite{dabre-etal-2022-indicbart,gala2023indictrans}\footnote{ai4bharat/IndicBART} vocabulary. To the best of our knowledge, there is no open-access generative model which understands both table structure and Bengali. We train the following models on \ac{BanglaTabQA} as they support Bengali and Hindi numbers and text:
\begin{enumerate}[label=(\arabic*),leftmargin=*,nosep]
\item \textbf{BnTQA-mBart:} \texttt{mbart-50-large}~\cite{liu-etal-2020-multilingual-denoising} is a multi-lingual encoder-decoder model with support for 50 languages.
\item \textbf{BnTQA-M2M:} \texttt{m2m100\_418M}~\cite{10.5555/3546258.3546365} is a multi-lingual encoder-decoder model with support for 100 languages.
\item \textbf{BnTQA-llama:} We train \texttt{Llama-7B}, on \ac{BanglaTabQA} dataset with \ac{PEFT} on LoRA adapters.
\end{enumerate}
We train \texttt{BnTQA-mBart} and \texttt{BnTQA-M2M} with 128 batch size and  \texttt{BnTQA-llama} with 16 batch size and 4-bit quantization. All models are trained with 1e-4 learning rate on a single A6000 48GB GPU for 5 epochs with 1024 maximum sequence length.

\subsection{HindiTabQA}
We assess the generalizabiltiy of our data generation process by training and evaluating HindiTabQA models. All hyper-parameters and experimental setup are the same as Bengali.

\paragraph{Baselines.} We use the following baselines:
\begin{enumerate}[label=(\arabic*),leftmargin=*,nosep]
\item \textbf{En2Hi:} Similar to \texttt{En2Bn}, we fine-tune \texttt{mbart-50-large} with $25,000$ random samples from MultiTabQA, translated to Hindi.
\item \textbf{GPT}: We perform 2-shot \ac{ICL} on the best LLMs on Bengali, \texttt{GPT-3.5} and \texttt{GPT-4}. 
\item \textbf{OpHathi}:  We perform 2-shot \ac{ICL} on \texttt{OpenHathi-7B-Hi-v0.1-Base}, an open-source LLM based on \texttt{llama-7b} and trained on Hindi, English, and Hinglish text.
\end{enumerate}

\paragraph{HindiTabQA models.} We train the following models on the HindiTabQA dataset:
\begin{enumerate}[label=(\arabic*),leftmargin=*,nosep]
\item \textbf{HiTQA-llama:} Similar to Bengali, we fine-tune \texttt{Llama-7b} on HindiTabQA dataset. 
\item \textbf{HiTQA-M2M:} Similar to Bengali, we fine-tune \texttt{m2m100\_418M} on HindiTabQA dataset.  
\item \textbf{HiTQA-mBart:} Similar to Bengali, we fine-tune \texttt{mbart-50-large}, on HindiTabQA. 
\item \textbf{HiTQA-BnTQA:} \texttt{BnTQA-mBart}, trained on BanglaTabQA provides a warm start. We fine-tune it on HindiTabQA for better convergence.
\end{enumerate}

\subsection{Evaluation Metrics}
The answer table requires both table structure and content evaluation rendering standard text similarity metrics (Rouge, BLEU, etc.) inappropriate. We instead evaluate with \ac{tableQA} evaluation metrics~\cite{pal2023multitabqa}. Henceforth, F1 scores are the harmonic mean of the precision and recall scores.

\begin{enumerate}[label=(\arabic*),leftmargin=*,nosep]
\item \textbf{Table Exact Match Accuracy (Tab)} measures the percentage of generated answer which \emph{match exactly} to the target answer tables. 
\item \textbf{Row Exact Match F1 (Row)}: Row EM precision is the percentage of correctly predicted rows among all predicted rows. Row EM recall is the percentage of correctly predicted rows among all target rows.
\item \textbf{Column Exact Match F1 (Col)}:
Column EM precision is the percentage of correctly predicted columns and corresponding headers among all predicted columns. Column EM recall is the percentage of correctly predicted columns among all target columns.
\item \textbf{Cell Exact Match F1 (Cell)} is the most relaxed metric. Cell EM precision is the percentage of correctly generated cells among all predicted cells. Cell EM recall is the percentage of correctly predicted cells among all target cells.
\end{enumerate}

\section{Results}
\label{sec:results}
\textbf{Baselines.} As reported in Table~\ref{tab:results_all_lang}, \texttt{GPT-4} performs the best on our test set with a table EM accuracy of $26.83\%$. \texttt{GPT-3.5} under-performs \texttt{GPT-4} but is better than open-sourced \ac{LLM}s. Open-source \ac{LLM}s, \texttt{OdiaG} is pre-trained on Bengali text data but not on structured table data. The low accuracy of \texttt{OdiaG} ($0.69\%$) can be attributed to the models' lack of table understanding and table specific question which differs significantly from text-based tasks on which it has been pre-trained on as shown in examples in Appendix \ref{sec:qualitaive_analysis}. Baseline encoder-decoder model, \texttt{En2Bn}, fine-tuned on translated tableQA data, correctly generates $4.73\%$ of rows and cells and under-performs \texttt{OdiaG}, but is better than \texttt{TableLlama}. Although fine-tuning improves Bengali understanding, the low scores can be attributed to the erroneous translations of English tables in the MultiTabQA dataset which corroborate with \cite{minhas-etal-2022-xinfotabs} that table translation leads to error-propagation to down-stream QA task. Further, a lack of culture-specific tables in the MultiTabQA pre-training dataset leads to downgraded performance on topics in the BanglaTabQA test set. In conclusion, \texttt{GPT-4} is able to perform table reasoning in low-resourced Bengali, but is very expensive and closed-source, limiting it's accessibility and utility. \texttt{GPT-3.5}'s and all open-access baseline models' low scores demonstrates the need for both task and language adaptation with a large-scale dataset for training accessible open-source language models for low-resourced \ac{tableQA}.

\paragraph{BanglaTabQA models.} 
Parameter-efficient fine-tuned Llama models, \texttt{BnTQA-llama}, achieves comparable results to \texttt{GPT-3.5}. Table~\ref{tab:results_all_lang} shows that fine-tuned encode-decoder models, \texttt{BnTQA-mBart} and \texttt{BnTQA-M2M}, outperforms \texttt{GPT-4} on table exact match accuracy (EM) and column EM F1, but not for row and cell EM F1. This can be attributed to incorrect header generation of \texttt{GPT-4} reflecting in column and subsequently table EM scores. Apart from \texttt{GPT-4}, all other baseline models underperform \ac{BanglaTabQA} encoder-decoder models by a large margin on all metrics. \texttt{BnTQA-llama} overfits to the validation set, and does not generalize well to the test set. The low scores of \ac{PEFT} compared to full fine-tuning (FT) can be attributed to insufficient alignment of the frozen parameters of the backbone Llama model and sub-optimal tokenization of Bengali which has been observed in SentencePiece tokenizers in non-Latin languages~\cite{banerjee-bhattacharyya-2018-meaningless,chinese-llama-alpaca}. The results establishes the quality of the \ac{BanglaTabQA} dataset and its effectiveness in adapting neural models to \textit{both} language and table understanding.

\begin{table}[!t]
\centering
\setlength{\tabcolsep}{0.7mm}
\resizebox{1\columnwidth}{!}{
\begin{tabular}{l cccc cccc}
\toprule
\textbf{Model} & \multicolumn{4}{c}{No post-processing} & \multicolumn{4}{c}{With post-processing} 
\\
\cmidrule(r){2-5}
\cmidrule{6-9}
BnTQA & \textbf{Tab} &  \textbf{Row} &  \textbf{Col} &  \textbf{Cell} & \textbf{Tab} &  \textbf{Row} &  \textbf{Col} &  \textbf{Cell} \\
\midrule
 -llama & $0.00$ & $0.00$ & \phantom{0}$0.00$ & $0.26$ & \phantom{0}$5.74$ & $17.59$ & \phantom{0}$5.69$ & $15.49$ \\ 
 -mBart & $0.00$ & $8.70$ & $10.74$ & $8.70$ & $19.01$ & $20.74$ & $19.01$ & $20.74$ \\ 
 -M2M & $0.00$ & $0.00$ & \phantom{0}$0.00$ & $0.00$& $18.18$ & $35.80$ & $18.18$ & $35.80$ \\
\bottomrule
\end{tabular}
}
\caption{Zero-shot cross-lingual transfer scores of BnTQA models on Hindi test data.}
\label{tab:0_shot_cross_lingual}
\end{table}

\begin{table*}[!htb]
    \begin{minipage}{.45\linewidth}
    \small
    \centering
    \begin{tabular}{ll}
    \toprule
    \textbf{Operator class} & \textbf{Operations} \\
    \midrule
    arithmetic (A) & count, sum, average, max, min \\
    sorting (So) & ascending, descending  \\
    groupBy (G) & table column/row grouping  \\
    filtering (F) & where, having \\
    set (Se) & union, intersect, except \\
    logical (L) & and, or, not, in, not in, between \\
    \bottomrule
    \end{tabular}
    \caption{Classification of tableQA operations.}
    \label{tab:operator_class}
    \mbox{}\vspace*{2.7\baselineskip}
    \end{minipage}%
    \begin{minipage}{.55\linewidth}
      \centering
        \small
            \resizebox{1\columnwidth}{!}{
            \begin{tabular}{@{~}p{3mm}p{5mm}p{5mm}p{5mm}p{6mm}|
            p{5mm}p{5mm}p{5mm}p{5mm}}
            \toprule
            \multirow{3}{*}{\textbf{\makecell{Op}}} & \multicolumn{4}{c}{Bengali} & \multicolumn{4}{c}{Hindi}\\
            \cmidrule(r){2-5}
            \cmidrule{6-9}
             & \textbf{Tab} & \textbf{Row} & \textbf{Col} & \textbf{Cell}  & \textbf{Tab} & \textbf{Row} & \textbf{Col} & \textbf{Cell} \\
            \cmidrule(r){2-5}
            \cmidrule{6-9}
            A &  $39.66$ & $55.64$ & $39.67$ & $55.64$  & $35.06$ & $41.71$ & $35.07$ & $41.71$ \\
            So & $25.00$ & $25.00$ & $25.00$ & $25.00$ & $\textbf{39.05}$ & $\textbf{42.74}$ & $\textbf{39.05}$ & $\textbf{42.74}$ \\
            G & $\textbf{50.00}$ & $\textbf{76.92}$ & $\textbf{50.00}$ & $\textbf{76.92}$  & $33.33$ & $35.96$ & $33.33$ & $35.96$ \\
            F & $37.78$ & $35.86$ & $37.77$ & $35.86$ & $23.23$ & $26.35$ & $23.23$ & $21.67$
            \\
            Se  & $36.11$ & $49.10$ & $36.11$ & $49.10$ & $\phantom{0}5.00$ & $11.11$ & $\phantom{0}5.00$ & $11.11$ 
            \\ 
            L & $34.38$ & $13.23$ & $34.38$ & $13.23$ & $25.58$ & $27.38$ & $25.58$ & $27.38$\\ 
            \bottomrule
            \end{tabular}
            }
            \caption{XTQA-mBart test set scores (\%) on Operator Class (Op); X is a low-resourced language (Bn or Hi).}
            \label{tab:operator_results_all_lang}
    \end{minipage} 
\end{table*}

\paragraph{HindiTabQA models.}
We follow a similar experimental setup as discussed in Section \ref{sec:experimental_setup}. We report the results in Table \ref{tab:results_all_lang}. We observe that \texttt{HiTQA-BnTQA}, initialized with with \texttt{BnTQA-mbart}, outperforms all \ac{HindiTabQA} models and achieves a test score of $41.32\%$. Similar to \ac{BanglaTabQA}, \texttt{HiTQA-mBart} outperforms \texttt{HiTQA-M2M} with a table EM test score of $33.06\%$ and $28.93\%$ respectively. \texttt{HiTQA-llama} underperforms compared to the encoder-decoder models. All models trained on the HindiTabQA dataset outperform the two-shot in-context learning baseline models. The results follow a similar trend to \ac{BanglaTabQA} models and prove that our data generation process is generalizable and the HindiTabQA dataset is able to align neural models for tableQA task in Hindi.

\subsection{Zero-shot Cross-lingual Transfer}
\label{sec:zero_shot_cross_lingual_transfer}
We further study generalizability, by selecting the best performing language, Bengali, and evaluating the \ac{BanglaTabQA} models on Hindi test set in a zero-shot setting \emph{without} training on Hindi data. This setup allows us to study the cross-lingual transfer of \ac{BanglaTabQA} models to Hindi with a different script, and evaluate how well the models generalize to new out-of-distribution input tables. \ac{BanglaTabQA} models are able to perform table reasoning in Hindi indicating semantic information transfer across languages. We demonstrate some examples in the Appendix \ref{sec:appendix_zero_shot}. Table headers and numbers generated from math operations are often in Bengali instead of Hindi (Example \ref{example:0-shot_example_9}). Extractive questions are generated correctly (Example \ref{example:0-shot_example_10}). Table~\ref{tab:0_shot_cross_lingual} lists the zero-shot cross-lingual scores using the original predictions (named ``No Post-Processing'') of the \ac{BanglaTabQA} models on the Hindi test set defined in Section~\ref{sec:HindiTableQA}. Additionally, we perform post-processing of the predictions to translate the predicted tables' values to Hindi. As translating tables, composed of numbers and entities, with machine translation systems is unreliable \cite{minhas-etal-2022-xinfotabs}, we follow an automatic post-processing pipeline to transform predicted answer tables to Hindi. First, all lexical occurrence of Bengali digits in predictions are replaced with Hindi digits using a dictionary. Next, all lexical occurrence of SQL keyword in Bengali in the prediction headers are replaced with a Bengali-to-SQL keyword mapping and subsequently with a SQL-to-Hindi mapping described in Section~\ref{sec:dataset_method}. This fixes most of the Bengali presence in the predictions. Finally, we translate the predicted column names/values in Bengali to Hindi with Google translate. Table~\ref{tab:0_shot_cross_lingual} shows that post-processing increases the scores, demonstrating the generalizability of \ac{BanglaTabQA} models' table reasoning capabilities on out-of-domain Hindi tables with unseen cultural entities. This further demonstrates the quality and utility of the \ac{BanglaTabQA} dataset and our proposed data generation method and quality of the trained models. 




\subsection{Mathematical Operator classes}
\label{sec:operator_class_results}
We study how \ac{BanglaTabQA} and \ac{HindiTabQA} datasets aid in Bengali and Hindi numeracy and math understanding by evaluating \texttt{BnTQA-mBart} and \texttt{HiTQA-mBart} on $6$ categories of operator classes (Section \ref{sec:operator_classes}). We observe in Table~\ref{tab:operator_results_all_lang} that \texttt{BnTQA-mbart} performs best on \textit{groupBy (G)} operators with a table EM accuracy of $50.00\%$ and \texttt{HiTQA-mBart} on \textit{Sorting (So)} operators with a table EM accuracy of $39.05\%$. Both models are able to generalize to unseen tables in the respective languages' test sets. This affirms that \ac{BanglaTabQA} and \ac{HindiTabQA} dataset aids mathematics reasoning of the trained models and enhances numeracy understanding in the low-resourced language.
\section{Conclusion}
\label{sec:conclusion}
Our work introduces \ac{tableQA} for the low-resource languages. We propose a methodology for large-scale dataset development on limited budget and automatic quality control which can be applied over any low-resource language with a web-presence. We discuss in detail the application of the methodology with an Indic Language, Bengali, for which we release a large-scale dataset, \ac{BanglaTabQA}. We further demonstrate generalizability of the process with another language, Hindi. We assess the datasets' quality by effectively training different Bengali and Hindi \ac{tableQA} models and conducting various experiments on model efficacy. Our studies on different operator classes and zero-shot cross-lingual transfer demonstrate that models trained with our dataset generalize well to unseen tables. Our proposed methodology can promote further research in low-resource \ac{tableQA}, while our released dataset and models can be used to further explore \ac{tableQA} for Bengali and Hindi.

\section*{Limitations}
We design a scalable automatic \ac{tableQA} data generation method and apply it on with two low-resourced languages: Bengali and Hindi. We release two \ac{tableQA} datasets: \ac{BanglaTabQA} and HindiTabQA and several models as outcome. Our main results in Table~\ref{tab:results_all_lang} demonstrate successful adaptation of neural models to low-resourced \ac{tableQA} task. Our extensive experimentation on generalizability in Section~\ref{sec:zero_shot_cross_lingual_transfer} and~\ref{sec:operator_class_results}  shows that models trained on the \ac{BanglaTabQA} dataset performs well across all operator classes and generalize to unseen languages and tables, proving generalizability of the datasets and methodology. 

Our dataset methodology is generalizable, but it is limited to languages for which unlabelled tables are available online. For very-low resource languages with low web presence, our method has only limited impact. Also, we used SQUALL templates for query generation, which do not support multi-table operations or complex queries. We leave addressing these challenges to future work.

\section*{Ethical Considerations}
The task and models proposed in the paper is aimed at closing the gap of resource scarcity in low-resource languages. To do so, we have used existing open-source resources publicly available in the web under MIT, CC-BY-SA-3.0 and  MIT, CC-BY-SA-4.0 licenses. Our dataset is generated synthetically data and will be released under MIT, CC-BY-SA-4.0 license. 
Our synthetic samples use templates from the SQUALL dataset also released under MIT, CC-BY-SA-4.0 license. Our test data splits are manually annotated. We pay each annotator €13.27/hour for their efforts. Further, we have utilized Wikipedia tables from Huggingface Wikipedia dataset. Wikipedia tables contain information about named-entities, facts and events in the public domain. We do not use any user-specific or sensitive data and information. Our models are built over open-source encoder-decoder models and closed-source GPT-3.5. Our work did not explicitly handle any bias which exists in the aforementioned pre-trained models or Wikipedia.

\section*{Acknowledgements}
We thank Elsevier's Discovery Lab for their support throughout this project and funding this work.
This work was also supported by 
Dutch Research Council (NWO), under project numbers 024.004.022, NWA.1389.20.\-183, KICH3.LTP.20.006, and VI.Vidi.223.166, and the European Union's Horizon Europe program under grant agreement No 101070212.
All content represents the opinion of the authors, which is not necessarily shared or endorsed by their respective employers and/or sponsors.

\bibliography{anthology,custom}

\begin{thebibliography}{54}
\expandafter\ifx\csname natexlab\endcsname\relax\def\natexlab#1{#1}\fi

\bibitem[{Banerjee and Bhattacharyya(2018)}]{banerjee-bhattacharyya-2018-meaningless}
Tamali Banerjee and Pushpak Bhattacharyya. 2018.
\newblock \href {https://doi.org/10.18653/v1/W18-1207} {Meaningless yet meaningful: Morphology grounded subword-level {NMT}}.
\newblock In \emph{Proceedings of the Second Workshop on Subword/Character {LE}vel Models}, pages 55--60, New Orleans. Association for Computational Linguistics.

\bibitem[{Bhattacharjee et~al.(2022)Bhattacharjee, Hasan, Ahmad, Mubasshir, Islam, Iqbal, Rahman, and Shahriyar}]{bhattacharjee-etal-2022-banglabert}
Abhik Bhattacharjee, Tahmid Hasan, Wasi Ahmad, Kazi~Samin Mubasshir, Md~Saiful Islam, Anindya Iqbal, M.~Sohel Rahman, and Rifat Shahriyar. 2022.
\newblock \href {https://doi.org/10.18653/v1/2022.findings-naacl.98} {{B}angla{BERT}: Language model pretraining and benchmarks for low-resource language understanding evaluation in {B}angla}.
\newblock In \emph{Findings of the Association for Computational Linguistics: NAACL 2022}, pages 1318--1327, Seattle, United States. Association for Computational Linguistics.

\bibitem[{Brown et~al.(2020)Brown, Mann, Ryder, Subbiah, Kaplan, Dhariwal, Neelakantan, Shyam, Sastry, Askell, Agarwal, Herbert-Voss, Krueger, Henighan, Child, Ramesh, Ziegler, Wu, Winter, Hesse, Chen, Sigler, Litwin, Gray, Chess, Clark, Berner, McCandlish, Radford, Sutskever, and Amodei}]{NEURIPS2020_1457c0d6}
Tom Brown, Benjamin Mann, Nick Ryder, Melanie Subbiah, Jared~D Kaplan, Prafulla Dhariwal, Arvind Neelakantan, Pranav Shyam, Girish Sastry, Amanda Askell, Sandhini Agarwal, Ariel Herbert-Voss, Gretchen Krueger, Tom Henighan, Rewon Child, Aditya Ramesh, Daniel Ziegler, Jeffrey Wu, Clemens Winter, Chris Hesse, Mark Chen, Eric Sigler, Mateusz Litwin, Scott Gray, Benjamin Chess, Jack Clark, Christopher Berner, Sam McCandlish, Alec Radford, Ilya Sutskever, and Dario Amodei. 2020.
\newblock \href {https://proceedings.neurips.cc/paper_files/paper/2020/file/1457c0d6bfcb4967418bfb8ac142f64a-Paper.pdf} {Language models are few-shot learners}.
\newblock In \emph{Advances in Neural Information Processing Systems}, volume~33, pages 1877--1901. Curran Associates, Inc.

\bibitem[{Cheng et~al.(2022)Cheng, Dong, Jia, Wu, Han, Cheng, and Zhang}]{cheng-etal-2022-fortap}
Zhoujun Cheng, Haoyu Dong, Ran Jia, Pengfei Wu, Shi Han, Fan Cheng, and Dongmei Zhang. 2022.
\newblock \href {https://doi.org/10.18653/v1/2022.acl-long.82} {{FORTAP}: Using formulas for numerical-reasoning-aware table pretraining}.
\newblock In \emph{Proceedings of the 60th Annual Meeting of the Association for Computational Linguistics (Volume 1: Long Papers)}, pages 1150--1166, Dublin, Ireland. Association for Computational Linguistics.

\bibitem[{Chowdhery et~al.(2022)Chowdhery, Narang, Devlin, Bosma, Mishra, Roberts, Barham, Chung, Sutton, Gehrmann, Schuh, Shi, Tsvyashchenko, Maynez, Rao, Barnes, Tay, Shazeer, Prabhakaran, Reif, Du, Hutchinson, Pope, Bradbury, Austin, Isard, Gur-Ari, Yin, Duke, Levskaya, Ghemawat, Dev, Michalewski, Garcia, Misra, Robinson, Fedus, Zhou, Ippolito, Luan, Lim, Zoph, Spiridonov, Sepassi, Dohan, Agrawal, Omernick, Dai, Pillai, Pellat, Lewkowycz, Moreira, Child, Polozov, Lee, Zhou, Wang, Saeta, Diaz, Firat, Catasta, Wei, Meier-Hellstern, Eck, Dean, Petrov, and Fiedel}]{chowdhery2022palm}
Aakanksha Chowdhery, Sharan Narang, Jacob Devlin, Maarten Bosma, Gaurav Mishra, Adam Roberts, Paul Barham, Hyung~Won Chung, Charles Sutton, Sebastian Gehrmann, Parker Schuh, Kensen Shi, Sasha Tsvyashchenko, Joshua Maynez, Abhishek Rao, Parker Barnes, Yi~Tay, Noam Shazeer, Vinodkumar Prabhakaran, Emily Reif, Nan Du, Ben Hutchinson, Reiner Pope, James Bradbury, Jacob Austin, Michael Isard, Guy Gur-Ari, Pengcheng Yin, Toju Duke, Anselm Levskaya, Sanjay Ghemawat, Sunipa Dev, Henryk Michalewski, Xavier Garcia, Vedant Misra, Kevin Robinson, Liam Fedus, Denny Zhou, Daphne Ippolito, David Luan, Hyeontaek Lim, Barret Zoph, Alexander Spiridonov, Ryan Sepassi, David Dohan, Shivani Agrawal, Mark Omernick, Andrew~M. Dai, Thanumalayan~Sankaranarayana Pillai, Marie Pellat, Aitor Lewkowycz, Erica Moreira, Rewon Child, Oleksandr Polozov, Katherine Lee, Zongwei Zhou, Xuezhi Wang, Brennan Saeta, Mark Diaz, Orhan Firat, Michele Catasta, Jason Wei, Kathy Meier-Hellstern, Douglas Eck, Jeff Dean, Slav Petrov, and Noah Fiedel. 2022.
\newblock \href {http://arxiv.org/abs/2204.02311} {{PaLM}: Scaling language modeling with pathways}.

\bibitem[{Cui et~al.(2023)Cui, Yang, and Yao}]{chinese-llama-alpaca}
Yiming Cui, Ziqing Yang, and Xin Yao. 2023.
\newblock \href {https://arxiv.org/abs/2304.08177} {Efficient and effective text encoding for {Chinese} {LLaMA} and {Alpaca}}.
\newblock \emph{arXiv preprint arXiv:2304.08177}.

\bibitem[{Dabre et~al.(2022)Dabre, Shrotriya, Kunchukuttan, Puduppully, Khapra, and Kumar}]{dabre-etal-2022-indicbart}
Raj Dabre, Himani Shrotriya, Anoop Kunchukuttan, Ratish Puduppully, Mitesh Khapra, and Pratyush Kumar. 2022.
\newblock \href {https://doi.org/10.18653/v1/2022.findings-acl.145} {{I}ndic{BART}: A pre-trained model for indic natural language generation}.
\newblock In \emph{Findings of the Association for Computational Linguistics: ACL 2022}, pages 1849--1863, Dublin, Ireland. Association for Computational Linguistics.

\bibitem[{Dahl et~al.(1994)Dahl, Bates, Brown, Fisher, Hunicke-Smith, Pallett, Pao, Rudnicky, and Shriberg}]{dahl-etal-1994-expanding}
Deborah~A. Dahl, Madeleine Bates, Michael Brown, William Fisher, Kate Hunicke-Smith, David Pallett, Christine Pao, Alexander Rudnicky, and Elizabeth Shriberg. 1994.
\newblock \href {https://aclanthology.org/H94-1010} {Expanding the scope of the {ATIS} task: The {ATIS}-3 corpus}.
\newblock In \emph{{H}uman {L}anguage {T}echnology: Proceedings of a Workshop held at {P}lainsboro, {N}ew {J}ersey, {M}arch 8-11, 1994}.

\bibitem[{Das and Saha(2022)}]{10.1007/s11042-021-11228-w}
Arijit Das and Diganta Saha. 2022.
\newblock \href {https://doi.org/10.1007/s11042-021-11228-w} {Deep learning based bengali question answering system using semantic textual similarity}.
\newblock \emph{Multimedia Tools Appl.}, 81(1):589–613.

\bibitem[{Deldjoo et~al.(2021)Deldjoo, Trippas, and Zamani}]{10.1145/3404835.3462806}
Yashar Deldjoo, Johanne~R. Trippas, and Hamed Zamani. 2021.
\newblock \href {https://doi.org/10.1145/3404835.3462806} {Towards multi-modal conversational information seeking}.
\newblock In \emph{Proceedings of the 44th International ACM SIGIR Conference on Research and Development in Information Retrieval}, SIGIR '21, page 1577–1587, New York, NY, USA. Association for Computing Machinery.

\bibitem[{Fan et~al.(2021)Fan, Bhosale, Schwenk, Ma, El-Kishky, Goyal, Baines, Celebi, Wenzek, Chaudhary, Goyal, Birch, Liptchinsky, Edunov, Grave, Auli, and Joulin}]{10.5555/3546258.3546365}
Angela Fan, Shruti Bhosale, Holger Schwenk, Zhiyi Ma, Ahmed El-Kishky, Siddharth Goyal, Mandeep Baines, Onur Celebi, Guillaume Wenzek, Vishrav Chaudhary, Naman Goyal, Tom Birch, Vitaliy Liptchinsky, Sergey Edunov, Edouard Grave, Michael Auli, and Armand Joulin. 2021.
\newblock Beyond {English}-centric multilingual machine translation.
\newblock \emph{J. Mach. Learn. Res.}, 22(1).

\bibitem[{Feng et~al.(2022)Feng, Yang, Cer, Arivazhagan, and Wang}]{2022_labse}
Fangxiaoyu Feng, Yinfei Yang, Daniel Cer, Naveen Arivazhagan, and Wei Wang. 2022.
\newblock \href {https://doi.org/10.18653/v1/2022.acl-long.62} {Language-agnostic bert sentence embedding}.
\newblock \emph{Proceedings of the 60th Annual Meeting of the Association for Computational Linguistics (Volume 1: Long Papers)}.

\bibitem[{Fleiss(1971)}]{Fleiss1971MeasuringNS}
Joseph~L. Fleiss. 1971.
\newblock \href {https://api.semanticscholar.org/CorpusID:143544759} {Measuring nominal scale agreement among many raters}.
\newblock \emph{Psychological Bulletin}, 76:378--382.

\bibitem[{Gala et~al.(2023)Gala, Chitale, Raghavan, Gumma, Doddapaneni, M, Nawale, Sujatha, Puduppully, Raghavan, Kumar, Khapra, Dabre, and Kunchukuttan}]{gala2023indictrans}
Jay Gala, Pranjal~A Chitale, A~K Raghavan, Varun Gumma, Sumanth Doddapaneni, Aswanth~Kumar M, Janki~Atul Nawale, Anupama Sujatha, Ratish Puduppully, Vivek Raghavan, Pratyush Kumar, Mitesh~M Khapra, Raj Dabre, and Anoop Kunchukuttan. 2023.
\newblock \href {https://openreview.net/forum?id=vfT4YuzAYA} {{IndicTrans2}: Towards high-quality and accessible machine translation models for all 22 scheduled {Indian} languages}.
\newblock \emph{Transactions on Machine Learning Research}.

\bibitem[{Gupta et~al.(2023)Gupta, Kandoi, Vora, Zhang, He, Reinanda, and Srikumar}]{gupta-etal-2023-temptabqa}
Vivek Gupta, Pranshu Kandoi, Mahek Vora, Shuo Zhang, Yujie He, Ridho Reinanda, and Vivek Srikumar. 2023.
\newblock \href {https://doi.org/10.18653/v1/2023.emnlp-main.149} {{T}emp{T}ab{QA}: Temporal question answering for semi-structured tables}.
\newblock In \emph{Proceedings of the 2023 Conference on Empirical Methods in Natural Language Processing}, pages 2431--2453, Singapore. Association for Computational Linguistics.

\bibitem[{Hasan et~al.(2020)Hasan, Bhattacharjee, Samin, Hasan, Basak, Rahman, and Shahriyar}]{hasan-etal-2020-low}
Tahmid Hasan, Abhik Bhattacharjee, Kazi Samin, Masum Hasan, Madhusudan Basak, M.~Sohel Rahman, and Rifat Shahriyar. 2020.
\newblock \href {https://doi.org/10.18653/v1/2020.emnlp-main.207} {Not low-resource anymore: Aligner ensembling, batch filtering, and new datasets for {B}engali-{E}nglish machine translation}.
\newblock In \emph{Proceedings of the 2020 Conference on Empirical Methods in Natural Language Processing (EMNLP)}, pages 2612--2623, Online. Association for Computational Linguistics.

\bibitem[{Hedderich et~al.(2021)Hedderich, Lange, Adel, Str{\"o}tgen, and Klakow}]{hedderich-etal-2021-survey}
Michael~A. Hedderich, Lukas Lange, Heike Adel, Jannik Str{\"o}tgen, and Dietrich Klakow. 2021.
\newblock \href {https://doi.org/10.18653/v1/2021.naacl-main.201} {A survey on recent approaches for natural language processing in low-resource scenarios}.
\newblock In \emph{Proceedings of the 2021 Conference of the North American Chapter of the Association for Computational Linguistics: Human Language Technologies}, pages 2545--2568, Online. Association for Computational Linguistics.

\bibitem[{Hendrycks et~al.(2021)Hendrycks, Burns, Basart, Zou, Mazeika, Song, and Steinhardt}]{hendryckstest2021}
Dan Hendrycks, Collin Burns, Steven Basart, Andy Zou, Mantas Mazeika, Dawn Song, and Jacob Steinhardt. 2021.
\newblock Measuring massive multitask language understanding.
\newblock \emph{Proceedings of the International Conference on Learning Representations (ICLR)}.

\bibitem[{Herzig et~al.(2020)Herzig, Nowak, M{\"u}ller, Piccinno, and Eisenschlos}]{herzig-etal-2020-tapas}
Jonathan Herzig, Pawel~Krzysztof Nowak, Thomas M{\"u}ller, Francesco Piccinno, and Julian Eisenschlos. 2020.
\newblock \href {https://doi.org/10.18653/v1/2020.acl-main.398} {{T}a{P}as: Weakly supervised table parsing via pre-training}.
\newblock In \emph{Proceedings of the 58th Annual Meeting of the Association for Computational Linguistics}, pages 4320--4333, Online. Association for Computational Linguistics.

\bibitem[{Hoffmann et~al.(2022)Hoffmann, Borgeaud, Mensch, Buchatskaya, Cai, Rutherford, de~Las~Casas, Hendricks, Welbl, Clark, Hennigan, Noland, Millican, van~den Driessche, Damoc, Guy, Osindero, Simonyan, Elsen, Rae, Vinyals, and Sifre}]{hoffmann2022training}
Jordan Hoffmann, Sebastian Borgeaud, Arthur Mensch, Elena Buchatskaya, Trevor Cai, Eliza Rutherford, Diego de~Las~Casas, Lisa~Anne Hendricks, Johannes Welbl, Aidan Clark, Tom Hennigan, Eric Noland, Katie Millican, George van~den Driessche, Bogdan Damoc, Aurelia Guy, Simon Osindero, Karen Simonyan, Erich Elsen, Jack~W. Rae, Oriol Vinyals, and Laurent Sifre. 2022.
\newblock \href {http://arxiv.org/abs/2203.15556} {Training compute-optimal large language models}.

\bibitem[{Hu et~al.(2022)Hu, Shen, Wallis, Allen-Zhu, Li, Wang, Wang, and Chen}]{hu2022lora}
Edward~J Hu, Yelong Shen, Phillip Wallis, Zeyuan Allen-Zhu, Yuanzhi Li, Shean Wang, Lu~Wang, and Weizhu Chen. 2022.
\newblock \href {https://openreview.net/forum?id=nZeVKeeFYf9} {Lo{RA}: Low-rank adaptation of large language models}.
\newblock In \emph{International Conference on Learning Representations}.

\bibitem[{Iyyer et~al.(2017)Iyyer, Yih, and Chang}]{iyyer-etal-2017-search}
Mohit Iyyer, Wen-tau Yih, and Ming-Wei Chang. 2017.
\newblock \href {https://doi.org/10.18653/v1/P17-1167} {Search-based neural structured learning for sequential question answering}.
\newblock In \emph{Proceedings of the 55th Annual Meeting of the Association for Computational Linguistics (Volume 1: Long Papers)}, pages 1821--1831, Vancouver, Canada. Association for Computational Linguistics.

\bibitem[{Jauhar et~al.(2016)Jauhar, Turney, and Hovy}]{Jauhar2016TablesAS}
Sujay~Kumar Jauhar, Peter~D. Turney, and Eduard~H. Hovy. 2016.
\newblock Tables as semi-structured knowledge for question answering.
\newblock In \emph{Annual Meeting of the Association for Computational Linguistics}.

\bibitem[{Jiang et~al.(2022)Jiang, Mao, He, Neubig, and Chen}]{jiang-etal-2022-omnitab}
Zhengbao Jiang, Yi~Mao, Pengcheng He, Graham Neubig, and Weizhu Chen. 2022.
\newblock \href {https://doi.org/10.18653/v1/2022.naacl-main.68} {{O}mni{T}ab: Pretraining with natural and synthetic data for few-shot table-based question answering}.
\newblock In \emph{Proceedings of the 2022 Conference of the North American Chapter of the Association for Computational Linguistics: Human Language Technologies}, pages 932--942, Seattle, United States. Association for Computational Linguistics.

\bibitem[{Jin et~al.(2022)Jin, Siebert, Li, and Chen}]{https://doi.org/10.48550/arxiv.2207.05270}
Nengzheng Jin, Joanna Siebert, Dongfang Li, and Qingcai Chen. 2022.
\newblock \href {https://doi.org/10.48550/ARXIV.2207.05270} {A survey on table question answering: Recent advances}.

\bibitem[{Joshi et~al.(2020)Joshi, Santy, Budhiraja, Bali, and Choudhury}]{joshi-etal-2020-state}
Pratik Joshi, Sebastin Santy, Amar Budhiraja, Kalika Bali, and Monojit Choudhury. 2020.
\newblock \href {https://doi.org/10.18653/v1/2020.acl-main.560} {The state and fate of linguistic diversity and inclusion in the {NLP} world}.
\newblock In \emph{Proceedings of the 58th Annual Meeting of the Association for Computational Linguistics}, pages 6282--6293, Online. Association for Computational Linguistics.

\bibitem[{Karim et~al.(2021)Karim, Kanti~Dey, Islam, Sarker, Hasan~Menon, Hossain, Hossain, and Decker}]{DeepHateExplainer}
Md.~Rezaul Karim, Sumon Kanti~Dey, Tanhim Islam, Sagor Sarker, Mehadi Hasan~Menon, Kabir Hossain, Md.~Azam Hossain, and Stefan Decker. 2021.
\newblock {DeepHateExplainer}: Explainable hate speech detection in under-resourced {Bengali} language.

\bibitem[{Katsis et~al.(2022)Katsis, Chemmengath, Kumar, Bharadwaj, Canim, Glass, Gliozzo, Pan, Sen, Sankaranarayanan, and Chakrabarti}]{aitqa2022}
Yannis Katsis, Saneem Chemmengath, Vishwajeet Kumar, Samarth Bharadwaj, Mustafa Canim, Michael Glass, Alfio Gliozzo, Feifei Pan, Jaydeep Sen, Karthik Sankaranarayanan, and Soumen Chakrabarti. 2022.
\newblock \href {https://doi.org/10.18653/v1/2022.naacl-industry.34} {{AIT-QA}: Question answering dataset over complex tables in the airline industry}.
\newblock \emph{Proceedings of the 2022 Conference of the North American Chapter of the Association for Computational Linguistics: Human Language Technologies: Industry Track}.

\bibitem[{Lin et~al.(2023)Lin, Blloshmi, Byrne, de~Gispert, and Iglesias}]{lin-etal-2023-inner}
Weizhe Lin, Rexhina Blloshmi, Bill Byrne, Adria de~Gispert, and Gonzalo Iglesias. 2023.
\newblock \href {https://doi.org/10.18653/v1/2023.acl-long.551} {An inner table retriever for robust table question answering}.
\newblock In \emph{Proceedings of the 61st Annual Meeting of the Association for Computational Linguistics (Volume 1: Long Papers)}, pages 9909--9926, Toronto, Canada. Association for Computational Linguistics.

\bibitem[{Liu et~al.(2021)Liu, Chen, Guo, Ziyadi, Lin, Chen, and guang Lou}]{liu2021tapex}
Qian Liu, Bei Chen, Jiaqi Guo, Morteza Ziyadi, Zeqi Lin, Weizhu Chen, and Jian guang Lou. 2021.
\newblock \href {http://arxiv.org/abs/2107.07653} {{TAPEX}: Table pre-training via learning a neural {SQL} executor}.

\bibitem[{Liu et~al.(2020)Liu, Gu, Goyal, Li, Edunov, Ghazvininejad, Lewis, and Zettlemoyer}]{liu-etal-2020-multilingual-denoising}
Yinhan Liu, Jiatao Gu, Naman Goyal, Xian Li, Sergey Edunov, Marjan Ghazvininejad, Mike Lewis, and Luke Zettlemoyer. 2020.
\newblock \href {https://doi.org/10.1162/tacl_a_00343} {Multilingual denoising pre-training for neural machine translation}.
\newblock \emph{Transactions of the Association for Computational Linguistics}, 8:726--742.

\bibitem[{Minhas et~al.(2022)Minhas, Shankhdhar, Gupta, Aggarwal, and Zhang}]{minhas-etal-2022-xinfotabs}
Bhavnick Minhas, Anant Shankhdhar, Vivek Gupta, Divyanshu Aggarwal, and Shuo Zhang. 2022.
\newblock \href {https://doi.org/10.18653/v1/2022.fever-1.7} {{XI}nfo{T}ab{S}: Evaluating multilingual tabular natural language inference}.
\newblock In \emph{Proceedings of the Fifth Fact Extraction and VERification Workshop (FEVER)}, pages 59--77, Dublin, Ireland. Association for Computational Linguistics.

\bibitem[{Nan et~al.(2022)Nan, Hsieh, Mao, Lin, Verma, Zhang, Kryściński, Schoelkopf, Kong, Tang, Mutuma, Rosand, Trindade, Bandaru, Cunningham, Xiong, and Radev}]{10.1162/tacl_a_00446}
Linyong Nan, Chiachun Hsieh, Ziming Mao, Xi~Victoria Lin, Neha Verma, Rui Zhang, Wojciech Kryściński, Hailey Schoelkopf, Riley Kong, Xiangru Tang, Mutethia Mutuma, Ben Rosand, Isabel Trindade, Renusree Bandaru, Jacob Cunningham, Caiming Xiong, and Dragomir Radev. 2022.
\newblock \href {https://doi.org/10.1162/tacl_a_00446} {{FeTaQA: Free-form Table Question Answering}}.
\newblock \emph{Transactions of the Association for Computational Linguistics}, 10:35--49.

\bibitem[{OpenAI et~al.(2023)OpenAI, Achiam, Adler, Agarwal, Ahmad, Akkaya, Aleman, Almeida, Altenschmidt, Altman, Anadkat, Avila, Babuschkin, Balaji, Balcom, Baltescu, Bao, Bavarian, Belgum, Bello, Berdine, Bernadett-Shapiro, Berner, Bogdonoff, Boiko, Boyd, Brakman, Brockman, Brooks, Brundage, Button, Cai, Campbell, Cann, Carey, Carlson, Carmichael, Chan, Chang, Chantzis, Chen, Chen, Chen, Chen, Chen, Chess, Cho, Chu, Chung, Cummings, Currier, Dai, Decareaux, Degry, Deutsch, Deville, Dhar, Dohan, Dowling, Dunning, Ecoffet, Eleti, Eloundou, Farhi, Fedus, Felix, Fishman, Forte, Fulford, Gao, Georges, Gibson, Goel, Gogineni, Goh, Gontijo-Lopes, Gordon, Grafstein, Gray, Greene, Gross, Gu, Guo, Hallacy, Han, Harris, He, Heaton, Heidecke, Hesse, Hickey, Hickey, Hoeschele, Houghton, Hsu, Hu, Hu, Huizinga, Jain, Jain, Jang, Jiang, Jiang, Jin, Jin, Jomoto, Jonn, Jun, Kaftan, Łukasz Kaiser, Kamali, Kanitscheider, Keskar, Khan, Kilpatrick, Kim, Kim, Kim, Kirchner, Kiros, Knight, Kokotajlo, Łukasz Kondraciuk,
  Kondrich, Konstantinidis, Kosic, Krueger, Kuo, Lampe, Lan, Lee, Leike, Leung, Levy, Li, Lim, Lin, Lin, Litwin, Lopez, Lowe, Lue, Makanju, Malfacini, Manning, Markov, Markovski, Martin, Mayer, Mayne, McGrew, McKinney, McLeavey, McMillan, McNeil, Medina, Mehta, Menick, Metz, Mishchenko, Mishkin, Monaco, Morikawa, Mossing, Mu, Murati, Murk, Mély, Nair, Nakano, Nayak, Neelakantan, Ngo, Noh, Ouyang, O'Keefe, Pachocki, Paino, Palermo, Pantuliano, Parascandolo, Parish, Parparita, Passos, Pavlov, Peng, Perelman, de~Avila Belbute~Peres, Petrov, de~Oliveira~Pinto, Michael, Pokorny, Pokrass, Pong, Powell, Power, Power, Proehl, Puri, Radford, Rae, Ramesh, Raymond, Real, Rimbach, Ross, Rotsted, Roussez, Ryder, Saltarelli, Sanders, Santurkar, Sastry, Schmidt, Schnurr, Schulman, Selsam, Sheppard, Sherbakov, Shieh, Shoker, Shyam, Sidor, Sigler, Simens, Sitkin, Slama, Sohl, Sokolowsky, Song, Staudacher, Such, Summers, Sutskever, Tang, Tezak, Thompson, Tillet, Tootoonchian, Tseng, Tuggle, Turley, Tworek, Uribe, Vallone,
  Vijayvergiya, Voss, Wainwright, Wang, Wang, Wang, Ward, Wei, Weinmann, Welihinda, Welinder, Weng, Weng, Wiethoff, Willner, Winter, Wolrich, Wong, Workman, Wu, Wu, Wu, Xiao, Xu, Yoo, Yu, Yuan, Zaremba, Zellers, Zhang, Zhang, Zhao, Zheng, Zhuang, Zhuk, and Zoph}]{openai2023gpt4}
OpenAI, Josh Achiam, Steven Adler, Sandhini Agarwal, Lama Ahmad, Ilge Akkaya, Florencia~Leoni Aleman, Diogo Almeida, Janko Altenschmidt, Sam Altman, Shyamal Anadkat, Red Avila, Igor Babuschkin, Suchir Balaji, Valerie Balcom, Paul Baltescu, Haiming Bao, Mo~Bavarian, Jeff Belgum, Irwan Bello, Jake Berdine, Gabriel Bernadett-Shapiro, Christopher Berner, Lenny Bogdonoff, Oleg Boiko, Madelaine Boyd, Anna-Luisa Brakman, Greg Brockman, Tim Brooks, Miles Brundage, Kevin Button, Trevor Cai, Rosie Campbell, Andrew Cann, Brittany Carey, Chelsea Carlson, Rory Carmichael, Brooke Chan, Che Chang, Fotis Chantzis, Derek Chen, Sully Chen, Ruby Chen, Jason Chen, Mark Chen, Ben Chess, Chester Cho, Casey Chu, Hyung~Won Chung, Dave Cummings, Jeremiah Currier, Yunxing Dai, Cory Decareaux, Thomas Degry, Noah Deutsch, Damien Deville, Arka Dhar, David Dohan, Steve Dowling, Sheila Dunning, Adrien Ecoffet, Atty Eleti, Tyna Eloundou, David Farhi, Liam Fedus, Niko Felix, Simón~Posada Fishman, Juston Forte, Isabella Fulford, Leo Gao,
  Elie Georges, Christian Gibson, Vik Goel, Tarun Gogineni, Gabriel Goh, Rapha Gontijo-Lopes, Jonathan Gordon, Morgan Grafstein, Scott Gray, Ryan Greene, Joshua Gross, Shixiang~Shane Gu, Yufei Guo, Chris Hallacy, Jesse Han, Jeff Harris, Yuchen He, Mike Heaton, Johannes Heidecke, Chris Hesse, Alan Hickey, Wade Hickey, Peter Hoeschele, Brandon Houghton, Kenny Hsu, Shengli Hu, Xin Hu, Joost Huizinga, Shantanu Jain, Shawn Jain, Joanne Jang, Angela Jiang, Roger Jiang, Haozhun Jin, Denny Jin, Shino Jomoto, Billie Jonn, Heewoo Jun, Tomer Kaftan, Łukasz Kaiser, Ali Kamali, Ingmar Kanitscheider, Nitish~Shirish Keskar, Tabarak Khan, Logan Kilpatrick, Jong~Wook Kim, Christina Kim, Yongjik Kim, Hendrik Kirchner, Jamie Kiros, Matt Knight, Daniel Kokotajlo, Łukasz Kondraciuk, Andrew Kondrich, Aris Konstantinidis, Kyle Kosic, Gretchen Krueger, Vishal Kuo, Michael Lampe, Ikai Lan, Teddy Lee, Jan Leike, Jade Leung, Daniel Levy, Chak~Ming Li, Rachel Lim, Molly Lin, Stephanie Lin, Mateusz Litwin, Theresa Lopez, Ryan Lowe,
  Patricia Lue, Anna Makanju, Kim Malfacini, Sam Manning, Todor Markov, Yaniv Markovski, Bianca Martin, Katie Mayer, Andrew Mayne, Bob McGrew, Scott~Mayer McKinney, Christine McLeavey, Paul McMillan, Jake McNeil, David Medina, Aalok Mehta, Jacob Menick, Luke Metz, Andrey Mishchenko, Pamela Mishkin, Vinnie Monaco, Evan Morikawa, Daniel Mossing, Tong Mu, Mira Murati, Oleg Murk, David Mély, Ashvin Nair, Reiichiro Nakano, Rajeev Nayak, Arvind Neelakantan, Richard Ngo, Hyeonwoo Noh, Long Ouyang, Cullen O'Keefe, Jakub Pachocki, Alex Paino, Joe Palermo, Ashley Pantuliano, Giambattista Parascandolo, Joel Parish, Emy Parparita, Alex Passos, Mikhail Pavlov, Andrew Peng, Adam Perelman, Filipe de~Avila Belbute~Peres, Michael Petrov, Henrique~Ponde de~Oliveira~Pinto, Michael, Pokorny, Michelle Pokrass, Vitchyr Pong, Tolly Powell, Alethea Power, Boris Power, Elizabeth Proehl, Raul Puri, Alec Radford, Jack Rae, Aditya Ramesh, Cameron Raymond, Francis Real, Kendra Rimbach, Carl Ross, Bob Rotsted, Henri Roussez, Nick Ryder,
  Mario Saltarelli, Ted Sanders, Shibani Santurkar, Girish Sastry, Heather Schmidt, David Schnurr, John Schulman, Daniel Selsam, Kyla Sheppard, Toki Sherbakov, Jessica Shieh, Sarah Shoker, Pranav Shyam, Szymon Sidor, Eric Sigler, Maddie Simens, Jordan Sitkin, Katarina Slama, Ian Sohl, Benjamin Sokolowsky, Yang Song, Natalie Staudacher, Felipe~Petroski Such, Natalie Summers, Ilya Sutskever, Jie Tang, Nikolas Tezak, Madeleine Thompson, Phil Tillet, Amin Tootoonchian, Elizabeth Tseng, Preston Tuggle, Nick Turley, Jerry Tworek, Juan Felipe~Cerón Uribe, Andrea Vallone, Arun Vijayvergiya, Chelsea Voss, Carroll Wainwright, Justin~Jay Wang, Alvin Wang, Ben Wang, Jonathan Ward, Jason Wei, CJ~Weinmann, Akila Welihinda, Peter Welinder, Jiayi Weng, Lilian Weng, Matt Wiethoff, Dave Willner, Clemens Winter, Samuel Wolrich, Hannah Wong, Lauren Workman, Sherwin Wu, Jeff Wu, Michael Wu, Kai Xiao, Tao Xu, Sarah Yoo, Kevin Yu, Qiming Yuan, Wojciech Zaremba, Rowan Zellers, Chong Zhang, Marvin Zhang, Shengjia Zhao, Tianhao
  Zheng, Juntang Zhuang, William Zhuk, and Barret Zoph. 2023.
\newblock \href {http://arxiv.org/abs/2303.08774} {Gpt-4 technical report}.

\bibitem[{Pal et~al.(2022)Pal, Kanoulas, and de~Rijke}]{pal-etal-2022-parameter}
Vaishali Pal, Evangelos Kanoulas, and Maarten de~Rijke. 2022.
\newblock \href {https://aclanthology.org/2022.dialdoc-1.5} {Parameter-efficient abstractive question answering over tables or text}.
\newblock In \emph{Proceedings of the Second DialDoc Workshop on Document-grounded Dialogue and Conversational Question Answering}, pages 41--53, Dublin, Ireland. Association for Computational Linguistics.

\bibitem[{Pal et~al.(2023)Pal, Yates, Kanoulas, and de~Rijke}]{pal2023multitabqa}
Vaishali Pal, Andrew Yates, Evangelos Kanoulas, and Maarten de~Rijke. 2023.
\newblock {MultiTabQA}: Generating tabular answers for multi-table question answering.
\newblock In \emph{ACL 2023: The 61st Annual Meeting of the Association for Computational Linguistics}, pages 6322--6634.

\bibitem[{Parida et~al.(2023)Parida, Sekhar, Kohli, Sen, and Sahoo}]{OdiaGenAI-Bengali-LLM}
Shantipriya Parida, Sambit Sekhar, Guneet~Singh Kohli, Arghyadeep Sen, and Shashikanta Sahoo. 2023.
\newblock Bengali instruction-tuning model.
\newblock \url{https://huggingface.co/OdiaGenAI}.

\bibitem[{Pasupat and Liang(2015)}]{pasupat-liang-2015-compositional}
Panupong Pasupat and Percy Liang. 2015.
\newblock \href {https://doi.org/10.3115/v1/P15-1142} {Compositional semantic parsing on semi-structured tables}.
\newblock In \emph{Proceedings of the 53rd Annual Meeting of the Association for Computational Linguistics and the 7th International Joint Conference on Natural Language Processing (Volume 1: Long Papers)}, pages 1470--1480, Beijing, China. Association for Computational Linguistics.

\bibitem[{Price(1990)}]{price-1990-evaluation}
Patti Price. 1990.
\newblock \href {https://aclanthology.org/H90-1020} {Evaluation of spoken language systems: the {ATIS} domain}.
\newblock In \emph{Speech and Natural Language: Proceedings of a Workshop Held at Hidden Valley, {P}ennsylvania, June 24-27,1990}.

\bibitem[{Ramesh et~al.(2022)Ramesh, Doddapaneni, Bheemaraj, Jobanputra, AK, Sharma, Sahoo, Diddee, J, Kakwani, Kumar, Pradeep, Nagaraj, Deepak, Raghavan, Kunchukuttan, Kumar, and Khapra}]{ramesh-etal-2022-samanantar}
Gowtham Ramesh, Sumanth Doddapaneni, Aravinth Bheemaraj, Mayank Jobanputra, Raghavan AK, Ajitesh Sharma, Sujit Sahoo, Harshita Diddee, Mahalakshmi J, Divyanshu Kakwani, Navneet Kumar, Aswin Pradeep, Srihari Nagaraj, Kumar Deepak, Vivek Raghavan, Anoop Kunchukuttan, Pratyush Kumar, and Mitesh~Shantadevi Khapra. 2022.
\newblock \href {https://doi.org/10.1162/tacl_a_00452} {Samanantar: The largest publicly available parallel corpora collection for 11 {I}ndic languages}.
\newblock \emph{Transactions of the Association for Computational Linguistics}, 10:145--162.

\bibitem[{Ruder(2019)}]{rudder-4-major-problems}
Sebastian Ruder. 2019.
\newblock The 4 biggest open problems in {NLP}.
\newblock \url{https://www.ruder.io/4-biggest-open-problems-in-nlp}.

\bibitem[{Shi et~al.(2020)Shi, Zhao, Boyd-Graber, {Daum\'{e} III}, and Lee}]{Shi:Zhao:Boyd-Graber:Daume-III:Lee-2020}
Tianze Shi, Chen Zhao, Jordan Boyd-Graber, Hal {Daum\'{e} III}, and Lillian Lee. 2020.
\newblock On the potential of lexico-logical alignments for semantic parsing to {SQL} queries.
\newblock In \emph{Findings of EMNLP}.

\bibitem[{Touvron et~al.(2023)Touvron, Martin, Stone, Albert, Almahairi, Babaei, Bashlykov, Batra, Bhargava, Bhosale, Bikel, Blecher, Ferrer, Chen, Cucurull, Esiobu, Fernandes, Fu, Fu, Fuller, Gao, Goswami, Goyal, Hartshorn, Hosseini, Hou, Inan, Kardas, Kerkez, Khabsa, Kloumann, Korenev, Koura, Lachaux, Lavril, Lee, Liskovich, Lu, Mao, Martinet, Mihaylov, Mishra, Molybog, Nie, Poulton, Reizenstein, Rungta, Saladi, Schelten, Silva, Smith, Subramanian, Tan, Tang, Taylor, Williams, Kuan, Xu, Yan, Zarov, Zhang, Fan, Kambadur, Narang, Rodriguez, Stojnic, Edunov, and Scialom}]{touvron2023llama}
Hugo Touvron, Louis Martin, Kevin Stone, Peter Albert, Amjad Almahairi, Yasmine Babaei, Nikolay Bashlykov, Soumya Batra, Prajjwal Bhargava, Shruti Bhosale, Dan Bikel, Lukas Blecher, Cristian~Canton Ferrer, Moya Chen, Guillem Cucurull, David Esiobu, Jude Fernandes, Jeremy Fu, Wenyin Fu, Brian Fuller, Cynthia Gao, Vedanuj Goswami, Naman Goyal, Anthony Hartshorn, Saghar Hosseini, Rui Hou, Hakan Inan, Marcin Kardas, Viktor Kerkez, Madian Khabsa, Isabel Kloumann, Artem Korenev, Punit~Singh Koura, Marie-Anne Lachaux, Thibaut Lavril, Jenya Lee, Diana Liskovich, Yinghai Lu, Yuning Mao, Xavier Martinet, Todor Mihaylov, Pushkar Mishra, Igor Molybog, Yixin Nie, Andrew Poulton, Jeremy Reizenstein, Rashi Rungta, Kalyan Saladi, Alan Schelten, Ruan Silva, Eric~Michael Smith, Ranjan Subramanian, Xiaoqing~Ellen Tan, Binh Tang, Ross Taylor, Adina Williams, Jian~Xiang Kuan, Puxin Xu, Zheng Yan, Iliyan Zarov, Yuchen Zhang, Angela Fan, Melanie Kambadur, Sharan Narang, Aurelien Rodriguez, Robert Stojnic, Sergey Edunov, and Thomas
  Scialom. 2023.
\newblock \href {http://arxiv.org/abs/2307.09288} {Llama 2: Open foundation and fine-tuned chat models}.

\bibitem[{Ye et~al.(2023)Ye, Hui, Yang, Li, Huang, and Li}]{10.1145/3539618.3591708}
Yunhu Ye, Binyuan Hui, Min Yang, Binhua Li, Fei Huang, and Yongbin Li. 2023.
\newblock \href {https://doi.org/10.1145/3539618.3591708} {Large language models are versatile decomposers: Decomposing evidence and questions for table-based reasoning}.
\newblock In \emph{Proceedings of the 46th International ACM SIGIR Conference on Research and Development in Information Retrieval}, SIGIR '23, page 174–184, New York, NY, USA. Association for Computing Machinery.

\bibitem[{Yin et~al.(2020)Yin, Neubig, Yih, and Riedel}]{tabert2020}
Pengcheng Yin, Graham Neubig, Wen-tau Yih, and Sebastian Riedel. 2020.
\newblock \href {https://doi.org/10.18653/v1/2020.acl-main.745} {{TaBERT}: Pretraining for joint understanding of textual and tabular data}.
\newblock \emph{Proceedings of the 58th Annual Meeting of the Association for Computational Linguistics}.

\bibitem[{Yu et~al.(2018)Yu, Zhang, Yang, Yasunaga, Wang, Li, Ma, Li, Yao, Roman, Zhang, and Radev}]{yu-etal-2018-spider}
Tao Yu, Rui Zhang, Kai Yang, Michihiro Yasunaga, Dongxu Wang, Zifan Li, James Ma, Irene Li, Qingning Yao, Shanelle Roman, Zilin Zhang, and Dragomir Radev. 2018.
\newblock \href {https://doi.org/10.18653/v1/D18-1425} {{S}pider: A large-scale human-labeled dataset for complex and cross-domain semantic parsing and text-to-{SQL} task}.
\newblock In \emph{Proceedings of the 2018 Conference on Empirical Methods in Natural Language Processing}, pages 3911--3921, Brussels, Belgium. Association for Computational Linguistics.

\bibitem[{Zelle and Mooney(1996)}]{10.5555/1864519.1864543}
John~M. Zelle and Raymond~J. Mooney. 1996.
\newblock Learning to parse database queries using inductive logic programming.
\newblock In \emph{Proceedings of the Thirteenth National Conference on Artificial Intelligence - Volume 2}, AAAI'96, page 1050–1055. AAAI Press.

\bibitem[{Zha et~al.(2023)Zha, Zhou, Li, Wang, Huang, Yang, Yuan, Su, Li, Su, Zhang, Zhou, Shou, Wang, Zhu, Lu, Ye, Ye, Ye, Zhang, Deng, Xu, Wang, Chen, and Zhao}]{zha2023tablegpt}
Liangyu Zha, Junlin Zhou, Liyao Li, Rui Wang, Qingyi Huang, Saisai Yang, Jing Yuan, Changbao Su, Xiang Li, Aofeng Su, Tao Zhang, Chen Zhou, Kaizhe Shou, Miao Wang, Wufang Zhu, Guoshan Lu, Chao Ye, Yali Ye, Wentao Ye, Yiming Zhang, Xinglong Deng, Jie Xu, Haobo Wang, Gang Chen, and Junbo Zhao. 2023.
\newblock \href {http://arxiv.org/abs/2307.08674} {{TableGPT}: Towards unifying tables, nature language and commands into one {GPT}}.

\bibitem[{Zhang et~al.(2024)Zhang, Pal, Huang, Kanoulas, and de~Rijke}]{zhang2024qfmtsgeneratingqueryfocusedsummaries}
Weijia Zhang, Vaishali Pal, Jia-Hong Huang, Evangelos Kanoulas, and Maarten de~Rijke. 2024.
\newblock \href {http://arxiv.org/abs/2405.05109} {Qfmts: Generating query-focused summaries over multi-table inputs}.

\bibitem[{Zhao et~al.(2022)Zhao, Li, Li, and Zhang}]{zhao-etal-2022-multihiertt}
Yilun Zhao, Yunxiang Li, Chenying Li, and Rui Zhang. 2022.
\newblock \href {https://doi.org/10.18653/v1/2022.acl-long.454} {{M}ulti{H}iertt: Numerical reasoning over multi hierarchical tabular and textual data}.
\newblock In \emph{Proceedings of the 60th Annual Meeting of the Association for Computational Linguistics (Volume 1: Long Papers)}, pages 6588--6600, Dublin, Ireland. Association for Computational Linguistics.

\bibitem[{Zhao et~al.(2023{\natexlab{a}})Zhao, Qi, Nan, Mi, Liu, Zou, Han, Chen, Tang, Xu, Radev, and Cohan}]{zhao-etal-2023-qtsumm}
Yilun Zhao, Zhenting Qi, Linyong Nan, Boyu Mi, Yixin Liu, Weijin Zou, Simeng Han, Ruizhe Chen, Xiangru Tang, Yumo Xu, Dragomir Radev, and Arman Cohan. 2023{\natexlab{a}}.
\newblock \href {https://doi.org/10.18653/v1/2023.emnlp-main.74} {{QTS}umm: Query-focused summarization over tabular data}.
\newblock In \emph{Proceedings of the 2023 Conference on Empirical Methods in Natural Language Processing}, pages 1157--1172, Singapore. Association for Computational Linguistics.

\bibitem[{Zhao et~al.(2023{\natexlab{b}})Zhao, Qi, Nan, Mi, Liu, Zou, Han, Tang, Xu, Cohan, and Radev}]{zhao2023qtsumm}
Yilun Zhao, Zhenting Qi, Linyong Nan, Boyu Mi, Yixin Liu, Weijin Zou, Simeng Han, Xiangru Tang, Yumo Xu, Arman Cohan, and Dragomir Radev. 2023{\natexlab{b}}.
\newblock \href {http://arxiv.org/abs/2305.14303} {{QTSumm}: A new benchmark for query-focused table summarization}.

\bibitem[{Zhong et~al.(2017)Zhong, Xiong, and Socher}]{zhong2017seq2sql}
Victor Zhong, Caiming Xiong, and Richard Socher. 2017.
\newblock \href {http://arxiv.org/abs/1709.00103} {{Seq2SQL}: Generating structured queries from natural language using reinforcement learning}.

\bibitem[{Zhu et~al.(2021)Zhu, Lei, Huang, Wang, Zhang, Lv, Feng, and Chua}]{zhu2021tat}
Fengbin Zhu, Wenqiang Lei, Youcheng Huang, Chao Wang, Shuo Zhang, Jiancheng Lv, Fuli Feng, and Tat-Seng Chua. 2021.
\newblock \href {https://doi.org/10.18653/v1/2021.acl-long.254} {{TAT}-{QA}: A question answering benchmark on a hybrid of tabular and textual content in finance}.
\newblock In \emph{Proceedings of the 59th Annual Meeting of the Association for Computational Linguistics and the 11th International Joint Conference on Natural Language Processing (Volume 1: Long Papers)}, pages 3277--3287, Online. Association for Computational Linguistics.

\end{thebibliography}
\bibliographystyle{acl_natbib}

\cleardoublepage
\appendix
\section{Appendix}
\label{sec:appendix}

\subsection{Bengali SQL2NQSim (LaBse fine-tuning) Results}
We evaluate semantic similarity of the LaBse model trained on the translated semantic parsing datasets comprising of Bengali SQL and it corresponding Bengali question (Section \ref{sec:nq_quality_control}) and report the validation set results in Table \ref{tab:SQL2NQSim_scores}. Both datasets show high semantic similarity among query-question pairs. However, \ac{BanglaTabQA} have a higher semantic similarity on various distance metrics indicating higher similarity of the query-question pairs compared to \ac{HindiTabQA}. \ac{HindiTabQA} lower semantic scores can be attributed to the lower recall scores among query-question pairs leading to lower F1 similarity scores. 
\begin{table}[h!]
    \centering
    \small
    \begin{tabular}{l@{}rr}
    \toprule
       \textbf{Scores} & \textbf{Bengali} & \textbf{Hindi} \\
       \midrule
         Accuracy with Cosine-Similarity & 91.99 & 98.67 \\
         F1 with Cosine-Similarity & 92.30 & 72.16  \\
         Precision with Cosine-Similarity & 94.55 & 77.68 \\
         Recall with Cosine-Similarity  & 90.15 & 67.36 \\
         Avg Precision with Cosine-Similarity & 97.79 & 75.32 \\
         Accuracy with Manhattan-Distance & 91.97 & 98.62 \\
         F1 with Manhattan-Distance & 92.31 & 70.96  \\
         Precision with Manhattan-Distance & 93.73 & 77.15 \\
         Recall with Manhattan-Distance & 90.94 & 65.69 \\
         Avg Precision with Manhattan-Distance & 97.80 & 74.41 \\
         Accuracy with Euclidean-Distance & 91.99 & 98.67 \\
         F1 with Euclidean-Distance & 92.30 & 72.16\\
         Precision with Euclidean-Distance & 94.55  & 77.68 \\
        Recall with Euclidean-Distance & 90.15 & 67.36 \\
        Avg Precision with Euclidean-Distance & 97.79 & 75.32 \\
        Accuracy with Dot-Product & 91.99 & 98.67\\
        F1 with Dot-Product & 92.30 & 72.16 \\
        Precision with Dot-Product & 94.55 & 77.68 \\
        Recall with Dot-Product & 90.15 & 67.36 \\
        Avg Precision with Dot-Product & 97.79 & 75.32\\
    \bottomrule
    \end{tabular}
    \caption{Bengali SQL2NQSim validation scores (\%)}
    \label{tab:SQL2NQSim_scores}
\end{table}

\subsection{Bengali SQL2NQ model Results}
We report the validation scores of the SQL2NQ models in Table \ref{tab:sql2nq_scores}. The Bengali SQL2NQ model scores are lower than the Hindi SQL2NQ model. Manual inspection of the generated dataset reveals that the Hindi questions and query have higher lexical overlap compared to the Bengali questions-query pairs where the questions are more natural leading to lower lexical overlap with the corresponding SQL query.
\begin{table}[h!]
    \small
    \centering
    \begin{tabular}{ccc}
    \toprule
        & \textbf{Bengali} & \textbf{Hindi} \\
        \midrule
        Rouge-1 & 14.63 & 53.20 \\
        Rouge-2 & \phantom{0}5.83 & 24.98 \\
        Rouge-L & 14.28 & 51.58 \\
    \bottomrule
    \end{tabular}
    \caption{Bengali SQL2NQ model's validation scores (\%)}
    \label{tab:sql2nq_scores}
\end{table}

\subsection{Open-Source Backbone Model Size}
We used the following open-source models as backbone to low-resource tableQA task. As observed in Table \ref{tab:model_sizes}, \texttt{M2M\_418} is the smallest backbone model among all models and \texttt{Llama-7b} is the largest.
\begin{table}
    \centering
    \small
    \begin{tabular}{lc}
    \toprule
    \textbf{Model} & \textbf{Number of Parameters} \\
    \midrule
       mbart-large-50  & 0.680 billion\\
       m2m100\_418M  & 0.418 billion \\
       Llama-7B   & \phantom{0.00}7 billion \\
       \bottomrule
    \end{tabular}
    \caption{Backbone model sizes}
    \label{tab:model_sizes}
\end{table}

\subsection{GPT Prompts} 
The 2-shot in-context learning prompt with demostrations to GPT is shown in Prompt~\ref{prompt:llm}: 

\begin{llmPrompt}[title={Prompt \thetcbcounter: 2-Shot ICL Prompt for GPT-3.5/4}, label=prompt:llm]
    {\bng Aapin Ekjn sHayk sHkarii iJin baNNGla pResh/nr Ut/tr edn baNNGla eTibl ethek baNNGlay Ut/tr eTibl oitir ker.} \small m {\bng sair EbNNG} \small n {\bng klamguilr EkiT eTibl inm/nilikht pYoaTaer/n elkha Hey:} <{\bng klam}> {\bng eTibl eHDar} <{\bng era 1}> {\bng man  1,1 . man 1,2 .} ... {\bng man 1,}\small n <{\bng era 2}> {\bng man 2,1 .} ... <{\bng era} \small m> {\bng man }\small m,{\bng1 . man }\small m,{\bng 2 .} ... {\bng . man} \small m,n

    \vspace{10pt}
    {\bng UdaHrN:}
    \vspace{10pt}

    {\bng 1)} \textbf{{\bng pRsh/n:}} {\bng kTa isheranam kaUn/TDaUn?} <{\bng klam}> {\bng  bchr . isheranam . bhuimka} <{\bng era 1}> \small 2006 {\bng . is ena Ibhl . ejkb guD naIT }...<{\bng era 13}> \small 2016 {\bng . kaUn/TDaUn . elh etRuainn }<{\bng era 14}> \small 2016 {\bng . kaUn/TDaUn . elh etRuainn} <{\bng era 15}> \small 2016 {\bng . kaUn/TDaUn . elh etRuainn}

    \vspace{10pt}
{\bng Ut/tr: }<{\bng klam}> {\bng gNna(`isherinam`) }<{\bng era 1}> {\bng 3}

\vspace{20pt}

{\bng 2) pRsh/n: kTa bcher isherinam sii ena Ibhl?} <{\bng klam}> {\bng  bchr . isheranam . bhuimka} <{\bng era 1}> \small 2006 {\bng . is ena Ibhl . ejkb guD naIT }<{\bng era 2}> \small 2006 {\bng . is ena Ibhl . ejkb guD naIT }<{\bng era 3}> \small 2006 {\bng . is ena Ibhl . ejkb guD naIT }...

\vspace{10pt}
{\bng Ut/tr:} <{\bng klam}> {\bng gNna(`bchr`)} <{\bng era 1}> {\bng  3}
\end{llmPrompt}

The English translation of the 2-shot prompt for in-context learning (ICL) of GPT-3.5/4 is shown in Prompt~\ref{prompt:llm_translation}:

\vspace*{-.5\baselineskip}
\begin{llmPrompt}[title={Prompt \thetcbcounter: 2-Shot ICL Prompt for GPT-3.5/4 (English translation)}, label=prompt:llm_translation]
    You are a helpful assistant who answers Bengali questions from Bengali tables by generating an answer table. A table of m rows and n columns is written in the following pattern: <column> table header <row 1> value 1,1 | value 1,2 | ... value 1,n <row 2> value 2,1 | ... <row m> value m,1 | value m,2 | ... | value m,n
    
    \textbf{Examples:}
    
    1) \textbf{Question:} How many titles are Countdown? <column> year | Title | Role <row 1> 2006 | See No Evil | Jacob Go ... <row 13>  2016 | Countdown  | Le Trunin <row 14>  2016 | Countdown  | Le Trunin  <row 15>  2016 | Countdown  | Le Trunin 

    \textbf{Answer:} <column> count(`Title`) <row 1> 3

    2) \textbf{Question:} How many years have See no Evil as titles? <column> year | Title | Role <row 1> 2006 | See No Evil | Jacob Good Night <row 2> 2006 | See No Evil | Jacob Good Night | <row 3> 2006 | See No Evil | Jacob Good Night ...  

    \textbf{Answer:} <column> count(`year`) <row 1> 3
\end{llmPrompt}

\subsection{LLama-based model Model Prompt}
The 2-shot in-context learning prompt with demostrations to Llama-7B based model, OdiaG,  is shown in Prompt ~\ref{prompt:llama}:
\label{sec:llama_prompt}
\begin{llmPrompt}[title={Prompt \thetcbcounter: 2-Shot ICL Prompt for odiagenAI-bn}, label=prompt:llama]
    \textbf{\#\#\# Instruction:}
    \begin{quote}
        {\bng Aapin Ekjn sHayk sHkarii iJin baNNGla eTibl oitir ker baNNGla pResh/nr Ut/tr edn.}
        {\bng UdaHrN:}
        
        \#\#\#Input:
        
        {\bng kTa iSeranam kaUn/TDaUn?} \small<{\bng klam}\small> {\bng bchr . iSeranam . bhuuimka} \small<{\bng era 1}> 2014 {\bng . sii ena Ebhl 2 . ejkb guD naIT} <{\bng era 2}> 2016 {\bng . kaUn/TDaUn . elh etRuinn} <{\bng era 3}> 2016 {\bng . kaUn/TDaUn . elh etRuinn}
         
        \#\#\# Response: 
        
         \small<{\bng klam}\small> {\bng gNna(iSeranam)} \small<{\bng era 1}> {\bng 2}
         
        \#\#\#End
        
        \#\#\#Input:
    
        {\bng kTa bchr iSeranam  sii ena Ebhl 2?} \small<{\bng klam}\small> {\bng bchr . iSeranam . bhuuimka} \small<{\bng era 1}> 2014 {\bng . sii ena Ebhl 2 . ejkb guD naIT} <{\bng era 2}> 2016 {\bng . kaUn/TDaUn . elh etRuinn} <{\bng era 3}> 2016 {\bng . kaUn/TDaUn . elh etRuinn}
         
        \#\#\# Response: 
        
         \small<{\bng klam}\small> {\bng gNna(iSeranam)} \small<{\bng era 1}> {\bng 1}
         
        \#\#\#End
        \vspace{1em}
        \end{quote}
        
        \textbf{\#\#\#Input:}
        
        \{input\}
        \vspace{1em} 
      
    \textbf{\#\#\# Response:}
\end{llmPrompt}

The English translation of the 2-shot in-context learning prompt with demostrations to Llama-7B based model, OdiaG, is shown in Prompt ~\ref{prompt:llama_translation}:

\begin{llmPrompt}[title={Prompt \thetcbcounter: 2-Shot ICL Prompt for odiagenAI-bn (English translation)}, label=prompt:llama_translation]
\textbf{\#\#\# Instruction:}
    \begin{quote}
            You are a helpful assistant who generates answers Bengali table to answer Bengali questions. Examples:

            \#\#\#Input: 
            
            How many titles are Countdown? <column> year | Title | Role <row 1>  2014 | See No Evil 2 | Jacob Goodnight <row 2>  2016 | Countdown  | Le Trunin <row 3>  2016 | Countdown  | Le Trunin 

            \#\#\#Response:  
            
            <column> count(Title) <row 1> 2 

            \#\#\# End
            
            \#\#\#Input: 
            
            How many years have See no Evil as titles? <column> year | Title | Role <row 1>  2014 | See No Evil 2 | Jacob Goodnight <row 2>  2016 | Countdown  | Le Trunin <row 3>  2016 | Countdown  | Le Trunin 

            \#\#\# Response:  
            
            <column> count(year) <row 1> 1 
    \end{quote}
     \vspace{1em}
    \textbf{\#\#\#Input:}
     
    \{input\}

    \vspace{1em} 
    \textbf{\#\#\#Response:}
\end{llmPrompt}


\subsection{BnTabQA Models Qualitative analysis}
\label{sec:qualitaive_analysis}
We analyze the output of each model with an example to identify error patterns and factors that impact model predictions. The test set question {\bng kar naem phuTsal smn/bykarii Athba pRJuit/tgt pircalekr Abs/than Aaech?} (Who has the position of Futsal Coordinator or Technical Director?), involves logical operator \texttt{or} after extracting values for {\bng phuTsal smn/bykarii} (Fulsal Coordinator) and {\bng pRJuit/tgt pircalekr} (Technical Director) from the column {\bng Abs/than} (\texttt{Position}). The input table is shown in Table~\ref{tab:input_bn_example} (translation of each table cell is italicized and in parenthesis for non-native readers) with target (English translation italicized and in parenthesis):
\begin{center}
{
\begin{tabular}{ll}
\hline
\makecell[l]{{\bng nam} (\emph{Name)})} \\
\hline
{\bng maIekl skubala} (\emph{Michael Skubala)} \\
{\bng els irD} (\emph{Les Reed)} \\
\hline
\end{tabular}
}
\end{center}

\begin{table*}[!t]
\centering
\small
\begin{tabular}{ll}
\hline
{\bng Abs/than} \emph{(Position)} & {\bng nam} \emph{(Name)}  \\
\hline
 {\bng sbhapit} \emph{(Chairman)} &  {\bng egRg k/lar/k} \emph{(Greg Clark)} \\ 
{\bng sH-sbhapit} \emph{(Co-Chairman)} & {\bng eDibhD igl} \emph{(David Gil)} \\
{\bng sadharN sm/padk} \emph{(General Secretary)} & {\bng mar/k builNNGHYam} \emph{(Mark Bullingham)} \\
{\bng ekaShadhY } \emph{(Treasurer)} & {\bng mar/k baeras} \emph{(Mark Burroughs)}  \\
\makecell{{\bng gNmadhYm EbNNG eJageJag pircalk} \emph{(Media and Communications Director)}} & {\bng luIsa iphyan/s} \emph{(Louisa Fiennes)} \\
{\bng pRJuittgt pircalk } \emph{(Technical Director)} & {\bng els irD} \emph{(Les Reed)}  \\
{\bng phuTsal smn/bykarii} \emph{(Futsal Coordinator)} & {\bng maIekl s/kubala} \emph{(Michael Skubala)}  \\
{\bng jatiiy delr ekac (puruSh)} \emph{(National Team Coach (Male))} & {\bng gYaerth saUthegT} \emph{(Gareth Southgate)}  \\
{\bng jatiiy delr ekac (narii)} \emph{(National Team Coach (Female))} & {\bng iphl enibhl} \emph{(Phil Neville)}  \\
{\bng erphair smn/bykarii} \emph{(Referee Coordinator)} & {\bng inl bYair} \emph{(Neil Barry)}  \\
\hline 
\end{tabular}

\caption{Example: BnTabQA Input Table. (English translation of each cell is italicized and in parenthesis)}
\label{tab:input_bn_example}
\end{table*}

%

\begin{example}\rm
Baseline encoder-decoder model, \texttt{En2Bn}, fine-tuned on the translated MultiTabQA dataset, correctly extracts {\bng maIekl s/kubala} (\emph{Michael Skubala)} as the {\bng phuTsal smn/bykarii} (Fulsal Coordinator), but wrongly assigns it as the table header instead of {\bng nam} (name). Moreover, it generates the same entity twice instead of generating {\bng els irD} (Les Reed):
\begin{center}
    \begin{tabular}{l}
        \hline
        \makecell[l]{{\bng phuTsal smn/bykarii} (\emph{Futsal Coordinator)}} 
        \\
        \hline
        {\bng maIekl skubala} (\emph{Michael Skubala)} \\
        {\bng maIekl skubala} (\emph{Michael Skubala)} \\
        \hline
    \end{tabular}
\end{center}
\end{example}

\begin{example}\rm
\texttt{OdiaG} also overfits to the demonstrations with {\bng gNna} (count) operator to generate incorrect value and header:
\begin{center}
    \begin{tabular}{l}
        \hline
        \makecell[l]{{\bng gNna(`nam`)} (\emph{count(Name)})} \\
        \hline
        \makecell[l]{{\bng 1} (\emph{1)}} \\
        \hline
    \end{tabular}
\end{center}
\end{example}

\begin{example}\rm
\texttt{GPT-3.5} with 2-shot in-context learning (ICL) extracts {\bng maIekl s/kubala} (\emph{Michael Skubala)} correctly but generates an incorrect table header over-fitting to the demonstrations:
\begin{center}
    \begin{tabular}{l}
        \hline
        \makecell[l]{{\bng gNna(`nam`)} (\emph{count(Name))}} \\
        \hline
        \makecell[l]{{\bng maIekl skubala} (\emph{Michael Skubala)}}
        \\
        \hline
    \end{tabular}
\end{center}
\end{example}

\begin{example}\rm
\texttt{GPT-4} with 2-shot in-context learning (ICL) correctly generates the answer table:
    \begin{center}
    \begin{tabular}{l}
        \hline
        \makecell[l]{{\bng nam} (\emph{Name)}} 
        \\
        \hline
        \makecell[l]{{\bng maIekl skubala} (\emph{Michael Skubala)}}
        \\
        \makecell[l]{{\bng els irD} (\emph{Les Reed)}} \\
        \hline
    \end{tabular}
\end{center}
\end{example}

\begin{example}\rm
Both encoder-decoder models, \texttt{BnTQA-mBart} and \texttt{BnTQA-M2M}, fine-tuned on BanglaTabQA dataset, correctly generates both answer table headers and values:
\begin{center}
    \begin{tabular}{l}
        \hline
        \makecell[l]{{\bng nam} (\emph{Name)}} \\
        \hline
        \makecell[l]{{\bng maIekl s/kubala} (\emph{Michael Skubala)}} \\
        \makecell[l]{{\bng els irD} (\emph{Les Reed)}} \\
        \hline
    \end{tabular}
\end{center}    
\end{example}


\begin{example}\rm
\texttt{BnTQA-Llama}, fine-tuned on BanglaTabQA dataset, is partially correct in its predictions by generating {\bng phuTsal smn/bykarii} (Fulsal Coordinator) in the first row, but incorrectly repeats the same entity instead of {\bng els irD} (\emph{Les Reed)} in the second row:
\begin{center}
    \begin{tabular}{l}
        \hline
        \makecell[l]{{\bng nam} (\emph{Name)}} \\
        \hline
        {\bng phuTsal smn/bykarii} (\emph{Fulsal Coordinator)} \\
        {\bng phuTsal smn/bykarii} (\emph{Fulsal Coordinator)} \\
        \hline
    \end{tabular}
\end{center}
\end{example}
We observe from the examples that all baselines except \texttt{GPT-4} generate wrong table headers and overfits and mimics the demonstrations, showing a lack of understanding of table structure and reasoning. The \ac{BanglaTabQA} models perform table reasoning, reflecting the utility and quality of the large-scale \ac{BanglaTabQA} dataset.

\subsection{Zero-Shot Cross-Lingual Transfer Examples}
\label{sec:appendix_zero_shot}
\begin{center}
\begin{table*}[th!]
\centering
\small
\begin{tabular}{lll}
\toprule
{\dn vq\0} (\emph{year}) & {\dn fFq\0k} (\emph{Title})  & {\dn EkrdAr } (\emph{Character}) \\
\midrule
2005 & {\dn \325wlAiV\3DAwAn}  (\emph{Flight Plan}) 
& {\dn eErk} (\emph{Eric})  \\
... & ... & ...  \\
2011  & {\dn in VAim}  (\emph{In Time})   & {\dn h\?nrF h\4EmSVn}  (\emph{Henry Hamilton})  \\
2011  & {\dn in VAim}  (\emph{In Time})   & {\dn h\?nrF h\4EmSVn}  (\emph{Henry Hamilton})  \\
2011  & {\dn in VAim} (\emph{In Time})   & {\dn h\?nrF h\4EmSVn}  (\emph{Henry Hamilton})  \\
2011  & {\dn in VAim}  (\emph{In Time})   & {\dn h\?nrF h\4EmSVn}  (\emph{Henry Hamilton}) \\
... & ... & ... \\
2014  & {\dn -p\?s -V\?fn \rn{76}}  (\emph{Space Station 76})   & {\dn V\4X}  (\emph{Ted})   \\
... & ... & ...  \\
2014  & {\dn Ev\2Vs\0 V\?l}  (\emph{Winter's Tale})   & {\dn pFVr l\?k k\? EptA} (\emph{Peter Lake's Father})   \\
\bottomrule
\end{tabular}
\caption{Example: HiTabQA Input Table (English translation of each cell is italicized and in parenthesis)}
\label{tab:hindi_example_input_table}
\end{table*}
\end{center}

\begin{example}\rm
\label{example:0-shot_example_9}
    The Hindi question, {\dn vq\0 \rn{2011} m\?{\qva} Ektn\? fFq\0k h\4{\qva}{\rs ?\re}} (\emph{How many titles are there in year 2011?}), with Hindi input table, Table~\ref{tab:hindi_example_input_table} (English translation is italicized and in parenthesis) and target table:
    \begin{center}
    \begin{tabular}{l}
        \hline
        \makecell[l]{{\dn gZnA{\rs (\re}fFq\0k{\rs )\re}}
 \emph{(count(Title))}} \\
        \hline
        \makecell[l]{{\dn } \includegraphics[width=0.04\columnwidth]{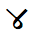} (\emph{4)}} \\
        \hline
    \end{tabular}
\end{center}
    \texttt{BnTQA-mBart} correctly performs table reasoning but generates the answer in Bengali script instead of Devnagari (Hindi) script:
    \begin{center}
    \begin{tabular}{l}
        \hline
        \makecell[l]{{\bng gnNa(isheranam) } \emph{(count(Title))}} \\
        \hline
        \makecell[l]{{\bng 4} (\emph{4)}} \\
        \hline
    \end{tabular}
\end{center}
\end{example}

\begin{example}\rm
\label{example:0-shot_example_10}
    However, for Hindi extractive questions like {\dn kOns\? \3FEwA\3D8wktA\0 aEDktm bAr aAy\? h\4{\qva}{\rs ?\re}}
 (Which recipient occurs the maximum number of times?), with Hindi input table:
    \begin{center}
        \begin{tabular}{ll}
            \toprule
            {\dn sAl} (\emph{year}) & {\dn \3FEwA\3D8wktA\0} (\emph{Recipient}) \\
            \midrule
            2016  & {\dn Evnod B\3D3w} (\emph{Vinod Bhatt})    \\
            2016  & {\dn Evnod B\3D3w} (\emph{Vinod Bhatt})    \\
            2017  & {\dn tArk mh\?tA{\rs [\re}\rn{1}{\rs ]\re}} (\emph{Tarak Mehta[1]})  \\
            \bottomrule
        \end{tabular}
    \end{center}
    and target table:
    \begin{center}
        \begin{tabular}{l}
            \hline
            \makecell[l]{{\dn \3FEwA\3D8wktA\0} \emph{(Recipient)}} \\
            \hline
            \makecell[l]{{\dn Evnod B\3D3w} (\emph{Vinod Bhatt)}} 
            \\
            \hline
        \end{tabular}
    \end{center}
    \texttt{BnTQA-mBart} correctly generates the answer in Hindi: 
    \begin{center}
        \begin{tabular}{l}
            \hline
            \makecell[l]{{\dn \3FEwA\3D8wktA\0} \emph{(Recipient)}} \\
            \hline
            \makecell[l]{{\dn Evnod B\3D3w} (\emph{Vinod Bhatt)}} \\
            \hline
        \end{tabular}
    \end{center}    
\end{example}


\subsection{Comparison of scores of LaBSE and SQL2NQ models}
\label{sec:labse_sql2nq_examples}
We qualitatively compare the sentence similarity models LaBse and SQL2NQ  with examples shown in Table \ref{tab:quality_control_examples}. We observe that LaBse scores are low for positive samples of Bengali SQL queries and the corresponding Bengali question. Further, negative samples, i.e., Bengali SQL query and an unrelated Bengali question has high similarity scores. This trend is not observed for the sentence similarity model, SQL2NQ, trained on Bengali SQL queries and corresponding Bengali natural questions.
\begin{table*}[!h]
\centering
\small
\resizebox{\textwidth}{!}{
\begin{tabular}{lllcc}
\toprule
 & & & \textbf{LaBse} & \textbf{SQL2NQ}
\\
 &\textbf{Bengali SQL} & \textbf{Bengali Question} & \textbf{Scores} & \textbf{Scores} \\
\midrule
\multirow{5}{*}{+ve} & \makecell[{{p{6.9cm}}}]{{\bng inr/bacon korun `bchr` dl kra `bchr` sajan eHak gnNa(`phlaphl`) siima 1} \\
(SELECT years GROUP BY years ORDER BY COUNT(result) LIMIT 1)} & \makecell[{{p{6.9cm}}}]{{\bng ekan bcher sbecey km phl Heyech?}\\
(Which year has the least number of results?)} & $0.45$ & $0.94$ \\
\cmidrule{2-5}
 &  \makecell[{{l}}]{{\bng inr/bacon korun `ishrnam` sajan eHak `bchr` AberaHii siima 1}  \\
(SELECT `title` ORDER BY `year` DESC LIMIT 1)} 
& \makecell[{{p{6.9cm}}}]{{\bng sm/pRitktm bcherr saeth sm/pRitk ishrnam ephrt idn.}\\
(Return the most recent title of the most recent year?)} & $0.43$ & $0.98$\\
\cmidrule{1-5}
\multirow{5}{*}{-ve} & \makecell[{{p{6.9cm}}}]{{\bng inr/bacon korun sr/binm/n(`sal`)} \\
(SELECT min(`year`))} & \makecell[{{p{6.9cm}}}]{{\bng ekan bcher (2010, 2016) sbecey ebtsh SNNGkhYk purSh/kar ijetech?}\\ (In which year (2010, 2016) were the most number of awards received?)} & $0.51$ & $0.31$ \\
\cmidrule{2-5}
&  \makecell[{{l}}]{{\bng inr/bacon korun gnNa(*) eJkhaen `kaj`="esoudaimniir sNNGsar"} \\
(SELECT count(*) WHERE `work`="The World of Saudamini")} & \makecell[{{p{6.9cm}}}]{{\bng eMaT 4 Aaech Emn egemr emaT SNNGkhYa gnNa krun.}\\ (How many games scored a total of 4?)} & $0.80$ & $0.07$ \\
\toprule
\end{tabular}
}
\caption{Comparison of sentence similarity scores between LaBse and our trained SQL2NQ models.}
\label{tab:quality_control_examples}
\end{table*}

\end{document}